\documentclass{article}

\usepackage{microtype}
\usepackage{graphicx}
\usepackage{subfigure}
\usepackage{booktabs} 
\usepackage{ulem}
\usepackage{hyperref}

 \usepackage[accepted]{icml2024}

\usepackage{amsmath}
\usepackage{amssymb}
\usepackage{mathtools}
\usepackage{amsthm}

\usepackage[capitalize,noabbrev]{cleveref}

\theoremstyle{plain}

\theoremstyle{definition}

\theoremstyle{remark}

\usepackage[textsize=tiny]{todonotes}

\icmltitlerunning{How Deep Networks Learn Sparse and Hierarchical Data: the Sparse Random Hierarchy Model}

\begin{document}

\twocolumn[
\icmltitle{How Deep Networks Learn Sparse and Hierarchical Data: \\
the Sparse Random Hierarchy Model}






\begin{icmlauthorlist}
\icmlauthor{Umberto M. Tomasini}{yyy}
\icmlauthor{Matthieu Wyart}{yyy}

\end{icmlauthorlist}

\icmlaffiliation{yyy}{Institute of Physics, EPFL, Lausanne, Switzerland}

\icmlcorrespondingauthor{Umberto, Tomasini}{umberto.tomasini@epfl.ch}

\icmlkeywords{Machine Learning, ICML}

\vskip 0.3in
]

\def\thefootnote{1}\footnotetext{Institute of Physics, EPFL, Lausanne, Switzerland. Correspondence to: Umberto Tomasini \textless \href{umberto.tomasini@epfl.ch}{umberto.tomasini@epfl.ch} \textgreater.\\

\textit{Proceedings of the 41 st International Conference on Machine
Learning}, Vienna, Austria. PMLR 235, 2024. Copyright 2024 by
the author(s).

}
\renewcommand{\thefootnote}{\arabic{footnote}}



\begin{abstract}
Understanding what makes high-dimensional data learnable is a fundamental question in machine learning. On the one hand, it is believed that the success of deep learning lies in its ability to build a hierarchy of representations that become increasingly more abstract with depth, going from simple features like edges to more complex concepts. On the other hand, learning to be insensitive to invariances of the task, such as smooth transformations for image datasets, has been argued to be important for deep networks and it strongly correlates with their performance. In this work, we aim to explain this correlation and unify these two viewpoints. We show that by introducing sparsity to generative hierarchical models of data, the task acquires insensitivity to spatial transformations that are discrete versions of smooth transformations. In particular, we introduce the Sparse Random Hierarchy Model (SRHM), where we observe and rationalize that a hierarchical representation mirroring the hierarchical model is learnt precisely when such insensitivity is learnt, thereby explaining the strong correlation between the latter and performance. Moreover, we quantify how the sample complexity of CNNs learning the SRHM depends on both the sparsity and hierarchical structure of the task.
\end{abstract}

\section{Introduction}
\label{sec:introduction}

Deep Learning has demonstrated remarkable efficacy across diverse tasks, from image classification \cite{voulodimos2018deep} to the development of chatbots \cite{brown2020language}. This success is not well understood: learning generic tasks requires a number of training points exponential in the data dimension \cite{luxburg2004distance,bach2017breaking}, which is unachievable in practice for images or text that lie in high dimension. Learnable data must then be highly structured. Understanding the nature of this structure is key to various fundamental problems of machine learning, ranging from the existence of scaling laws \cite{cortes1993learning,hestness2017deep,spigler2019asymptotic,kaplan2020scaling, bommasani2022opportunities} that relate the size of the training set to the performance and the emergence of new abilities,  the success \textcolor{black}{of transfer learning,} or that of unsupervised methods such as diffusion models \cite{ho2020denoising}.

One view is that learnable data often consists of local features that are assembled hierarchically \cite{grenander1996elements}: a dog is made of a body, limbs, and a face, itself consisting of ears, eyes, etc... This view is consistent with the observation that after training, deep neural networks form a hierarchical representation of the input \cite{zeiler_visualizing_2014,doimo2020hierarchical}. These properties also mirror the architecture of Convolutional Neural Networks (CNNs) \cite{lecun_gradient-based_1998,Lecun15}, which are deep and display small filter sizes. From a theoretical viewpoint, \textcolor{black}{the locality of the features proves to be advantageous} \textcolor{black}{ to approximate and learn high-dimensional tasks}
\cite{favero_locality_2021, Abbe21staircase,bietti2022learning,xiao2022eigenspace, pmlr-v162-xiao22a, bietti2022approximation,  mei2022generalization, cagnetta2023can}.
\textcolor{black}{Moreover, various hierarchical models of data, organizing local features within a hierarchical structure, have been proposed, } \cite{mossel2016deep, poggio2017why, malach2018provably, schmidt2020nonparametric, malach2020implications,cagnetta2023deep,allen-zhu2023how}. \textcolor{black}{As detailed in \autoref{sec:prior_work}, deep networks can represent and learn such models more efficiently than shallow networks, both in terms of number of parameters and number of training points.}

\begin{figure*}[h]
    \centering
    \includegraphics[width=1\textwidth]{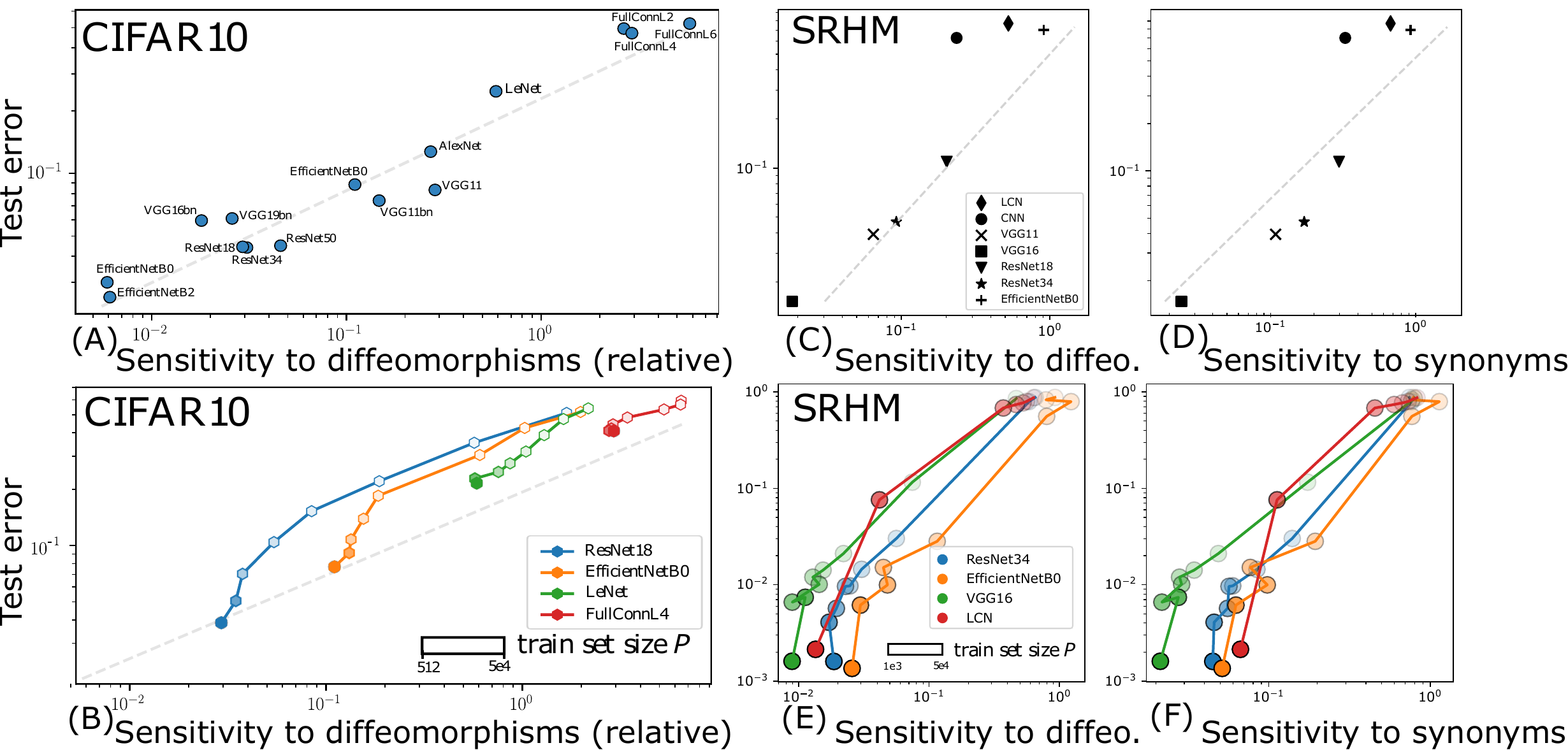}
    \caption{\textbf{CIFAR 10:} (A) Test error vs sensitivity to diffeomorphisms of common architectures trained on all CIFAR10, showing a remarkable correlation between the two quantities. A grey line, corresponding to a power-law, guides the eye. (B) Same as (A), for increasing size of the training set $P$, whose value is indicated by the degree of opacity. The sensitivity to smooth transformations is computed in relative terms to the sensitivity to white noise. (A, B) are adapted from \cite{petrini_relative_2021}.  \textbf{ The SRHM captures these observations:} (C) Test error vs sensitivity to diffeomorphisms of a CNN trained with $P=7400$  on the SHRM model, with parameters $L=s=s_0=2$ and $n_c=m=10$. The sensitivity to diffeomorphisms is defined as the change of network output induced by a diffeomorphism applied on the input, see Eq. \ref{eq:d_2}. For details about the architectures and their training process, see \autoref{app:sens_testerror_newnets}.  (D) Same as (C)  for sensitivity to exchange of synonyms, defined as the change of the network output induced by an exchange of synonyms \textcolor{black}{(defined in Section 2)} applied on the input, see Eq. \ref{eq:s_2}. (E) and (F): as top panels (C) and (D), for increasing $P$ (increasing opacity).}
   
    \label{fig:phasediagram_new}
\end{figure*}

Intrinsically, hierarchical models have a discrete nature, which seems well-suited for texts. Images however are often approximated as a continuous function of a two-dimensional space \cite{castleman1996digital}, whose label is invariant to smooth transformations.
This stability of image labels to small smooth transformations has been proposed in ~\cite{bruna2013invariant,mallat2016understanding} as a key simplification enabling image classification in high dimension. Enforcing such stability in neural networks can improve their performance \cite{kayhan2020translation}. Moreover, the stability can also improve during training by learning  pooling operations, \cite{dieleman2016exploiting, azulay2018deep, ruderman_pooling_2018, zhang2019making,Tomasini_2023}. 
One finds that (i) there exists a strong correlation between the network's test error and its sensitivity to diffeomorphisms, which characterizes the change of output when a diffeomorphism is applied to the input ~\cite{petrini_relative_2021}, as recalled in \autoref{fig:phasediagram_new} (A) and (ii)  the sensitivity to diffeomorphisms decreases with the size of the training set, as shown in \autoref{fig:phasediagram_new} (B). Currently, these observations remain unexplained, \textcolor{black}{and it is not clear how they relate with the fact that neural networks build a hierarchical representations of data.} 


\subsection{ Our contributions}
\begin{itemize}
\item We argue that incorporating sparsity into hierarchical generative models naturally leads to classification tasks \textcolor{black}{insensitive to the exact position of the local features, implying insensitivity to} discrete versions of diffeomorphisms.

\item \textcolor{black}{To illustrate this process, }we introduce the Sparse Random Hierarchy Model (SRHM), which captures the empirically observed correlation between sensitivity to \nolinebreak diffeomorphisms and test error,  as depicted in Fig.\ref{fig:phasediagram_new} \textcolor{black}{(C)}.

\item This correlation arises because a hierarchical representation, crucial for achieving good performance, is learnt precisely at the same number of training points at which insensitivity to diffeomorphisms is achieved. We provide arguments justifying this observation. 

\item We quantify how the \textcolor{black}{number of training points needed to learn the task, also called} sample complexity, of deep networks is influenced by both the sparsity and hierarchical nature of the task, and how it depends on the presence of weight sharing in the architecture. 
\end{itemize}

\begin{figure*}[h]
    \centering
    \includegraphics[width=.90\textwidth]{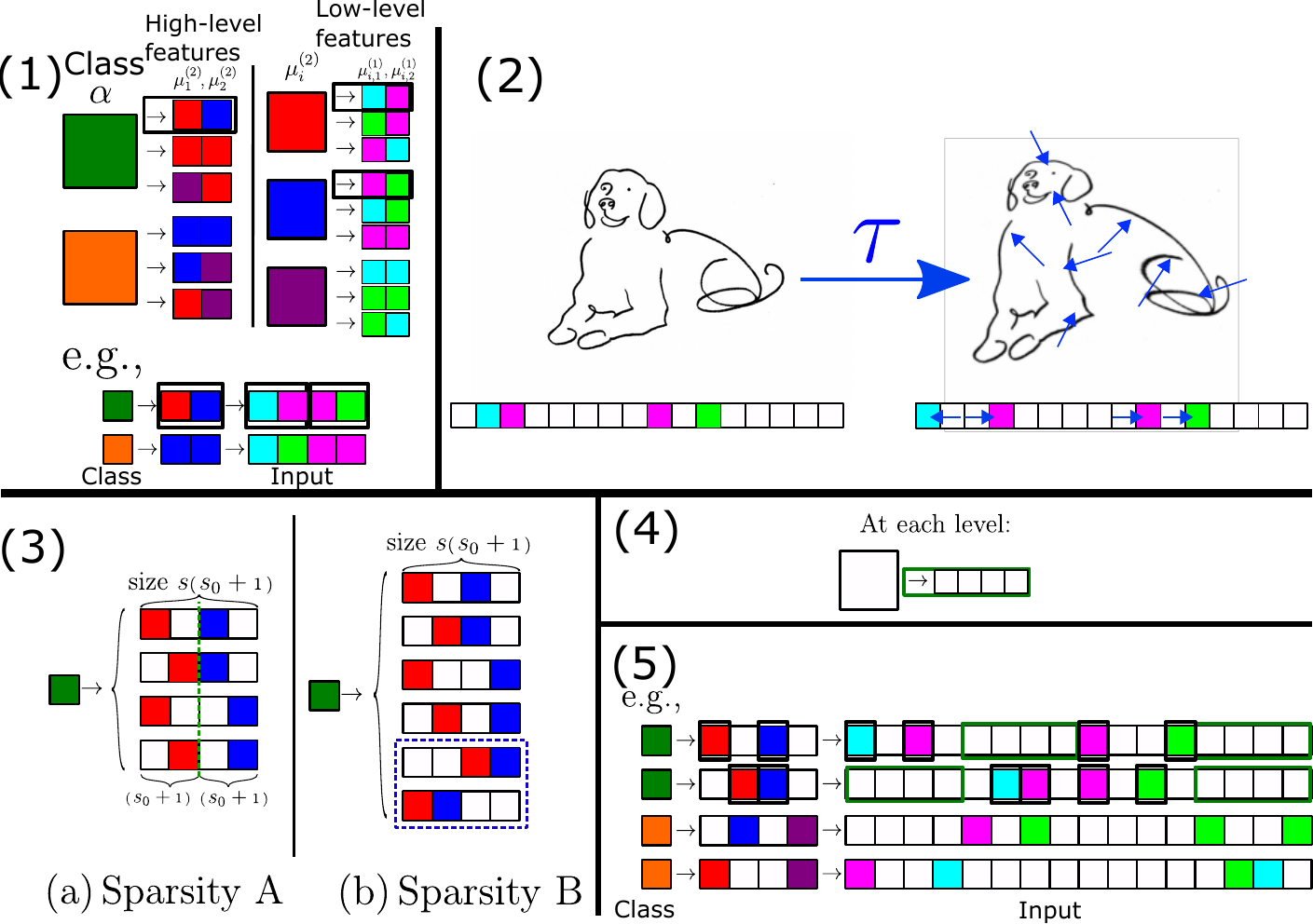}
    
    \caption{\textbf{(1)} \textcolor{black}{On the top of that panel,} an instance of the production rules of a Random Hierarchical Model (RHM) with $n_c=2$ classes, $L=2$,  $m=3$ synonyms per feature, vocabulary size $v=3$, and $s=2$. Here the classes set is $\mathcal{C}=\{\text{green, orange}\}$, the high-level features vocabulary is $\mathcal{V}_2=\{\text{red, blue, purple}\}$ and the low-level features vocabulary is $\mathcal{V}_1=\{\text{turquoise, pink, green}\}$. \textcolor{black}{On the bottom, a couple of examples generated via the rules above are shown. The first example is generated by the production rules in the black boxes (i.e. the green label generates (red,blue), which themselves generate the couples (turquoise, pink) and (pink, green). } \textbf{(2)} Top: effect of a diffeomorphism $\tau$ on a dog. The blue arrows represent the displacement field induced by $\tau$. Bottom: effect of a diffeomorphism $\tau$ on an instance of a sparse generative hierarchical task.  \textbf{(3)} Different definitions of sparsity. (A) Each one of the $s$ informative features is embedded in a sub-patch of size $s_0+1$ with strictly $s_0$ uninformative elements, yielding a patch of $s(s_0+1)$ elements. (B) The $s$ informative features can occupy any position within the patch of $s(s_0+1)$ elements. In both cases, all the possible rearrangements are shown for $s=2$ and $s_0=1$. At the next production rule, each uninformative element generates an empty patch of $s(s_0+1)$ uninformative elements, as pictured in \textbf{(4)}. \textbf{(5)} Four data sampled from the generative hierarchical task shown in panel (1) with sparsity (A). The first two examples follow the rules in black boxes in panels (1) and (4), showcasing different feature rearrangements.  }
    \label{fig:rhm_all}
\end{figure*}

\subsection{Prior work } \label{sec:prior_work}
{\it Hierarchical representations:} It is recognized that deep networks can represent hierarchically compositional functions with significantly fewer parameters than shallow networks \cite{poggio2017why}. Sufficient information exists in a training set with size polynomial in $d$ to reconstruct such tasks~\cite{schmidt2020nonparametric}--- although the training may take an immense time. Generative hierarchical models of data \cite{mossel2016deep, malach2018provably, malach2020implications} can be learnt efficiently by iterative clustering methods, if correlations exist between input features and output. Deep architectures trained with gradient descent display a hierarchical representation of the data, corresponding to the latent variables in these models  \cite{cagnetta2023deep,allen-zhu2023how}. Deep networks, as opposed to shallow ones, beat the curse of dimensionality, with a sample complexity that is polynomial in the dimension \cite{cagnetta2023deep}. However, none of these works consider sparsity in feature space, and the stability it confers to discretised smooth transformations of the input. 

{\it Task structure and invariance:}
\textcolor{black}{In \cite{hupkes21}  a classification of different combinatorial properties of tasks is introduced. The SRHM displays  the properties of systematicity, substitutivity and localism in that classification \cite{hupkes21}\looseness -1.}

\textcolor{black}{ CNNs were crafted to have internal representations equivariant to certain transformations such as rotations \cite{pmlr-v32-cohen14,pmlr-v48-cohenc16,pmlr-v76-ensign17a,pmlr-v80-kondor18a,pmlr-v139-finzi21a, batzner22,blumsmith2023machine}. How such a procedure can improve sample complexity
has been addressed for linear models such as random features and kernel methods  \cite{favero_locality_2021,bietti2021sample,pmlr-v139-elesedy21a,NEURIPS2021_8fe04df4_elesedyb,pmlr-v134-mei21a}.
Instead, our work focuses on how  invariance emerges during training and how it affects sample complexity, in a regime where features are learnt. } 

\textcolor{black}{Finally, in the context of adversarial robustness the response of trained networks to small-norm transformations of the input data that change the input label are investigated \cite{fawzi_manitest_2015,kanbak_geometric_2018,alaifari_adef_2018,athalye_synthesizing_2018,xiao_spatially_2018,alcorn_strike_2019,engstrom_exploring_2019}. Our approach differs from this literature since we consider the response to  generic perturbations belonging to specific ensembles, rather than worst-case perturbations.}


\section{Background: hierarchical generative models}
\label{sec:model}

\textcolor{black}{We consider} hierarchical classification tasks \textcolor{black}{where} images are generated from class labels through a hierarchical composition of production rules ~\cite{mossel2016deep,malach2018provably,degiuli2019random,malach2020implications, cagnetta2023deep,allen-zhu2023how}\textcolor{black}{. These tasks} represent a specific case of context-free grammars, a generative model in formal language theory  \cite{rozenberg_handbook_1997}.  
We focus on $L$-level context-free grammar, considering a set of class labels $\mathcal{C}\equiv\left\lbrace 1,\dots, n_c\right\rbrace$ and $L$ disjoint vocabularies $\mathcal{V}_\ell\,{\equiv}\,\left\lbrace a^{\ell}_1,\dots,a^{\ell}_{v_\ell}\right\rbrace$ of low- and high-level features, with $\ell=1,...,L$.  Henceforth, we refer to $\ell>1$ as high-level features or latent variables. Upon selecting a class label $\alpha$ uniformly at random from $\mathcal{C}$, the data is generated iteratively from a set of production rules: 
\begin{eqnarray}
\label{pro}\alpha \mapsto \mu^{(L)}_1,\dots,\mu^{(L)}_{s_{L}}\text{ for }\alpha\in\mathcal{C}\text{ and }\mu^{(L)}_i \in \mathcal{V}_{L},\\
\mu^{(\ell)} \mapsto \mu^{(\ell-1)}_{1},\dots,\mu^{(\ell-1)}_{s_{\ell}}\text{ for }\mu^{(\ell)}\in\mathcal{V}_{\ell},\mu^{(\ell-1)}_{i} \in \mathcal{V}_{\ell-1},
\label{production}
\end{eqnarray}
for $\ell=2,...,L$. At each level $\ell$, we consider  $m_{\ell}$ distinct rules stemming from each higher-level feature $a^{(\ell)}_i$. In other words, there are $m_{\ell}$ equivalent lower-level representations of $a^{(\ell)}_i$ for all $i=1,...,v$. 
These equivalent representations are termed `synonyms'. \textcolor{black}{ In \autoref{fig:rhm_all}, panel 1, we show on the top the generation rules of a generative hierarchical task with $n_c=2$ classes with two levels of production rules, each counting 3 synonyms.}

To simplify notation, we opt for the case of the Random Hierarchy Model (RHM) \cite{cagnetta2023deep}, for which: (i) $\forall \ell, v_{\ell}=v, m_{\ell}=m, s_{\ell}=s$, (ii) the $m$ production rules associated with any latent variable or class, \textcolor{black}{ shown on top in the example of panel 1 in \autoref{fig:rhm_all}}, are randomly chosen among the $v^s$ possibles ones and (iii) the $m$ production rules of any latent variable or class are sampled uniformly during data generation, \textcolor{black}{ as exemplified on the bottom of panel 1 in \autoref{fig:rhm_all}}. \textcolor{black}{Consequently}, the total number of possible data produced per class is $m^{\frac{d-1}{s-1}}$, where the dimension $d$ is defined as  $d=s^L$. (iv)  Two distinct classes or latent variables cannot yield the same low-level representation. This condition ensures, for example, that two distinct classes never generate the same data, implying that $m\leq v^{s-1}$. An example of a hierarchical generation process is shown in \autoref{fig:rhm_all}, panel (1).

Note that (i) we represent each input level feature in  $\mathcal{V}_1$ with a one-hot encoding of length $v$, yielding input data with dimension $d\times v$. The representation of higher-level features need not be specified, as they are latent variables.  (ii) Although we focused on generating one-dimensional data patches, the same model can generate square `images' without modifications if $s$ is the square of a natural number.

The RHM can be used to generate $P$ training points of input-label pairs $(x,y)$ where $y=1,..,n_c$ indicates the class and $x\in \mathbb{R}^{d\times v}$ is the input corresponding to $d$ sub-features $\mu^{(1)}$, each one-hot encoded in a dimension $v$. Using such data to train deep networks to classify $y$ from $x$, one finds  \cite{cagnetta2023deep} that:
\begin{itemize}
    \item The sample complexity $P^*$ of shallow network grows exponentially in $d$, whereas for deep networks it is polynomial in $d$ and reads $P^*\propto  n_c m^L$, with a prefactor of order unity for CNNs. 
    
    \item The sample complexity corresponds also to the training set size at which a hierarchical neural representation emerges. The latent variables $\mu^\ell$ are encoded closer to the output as $\ell$ increases. This representation is insensitive to change of synonyms in the input.
    
    \item Synonyms produced by the same high-level feature $\mu^{(2)}$ of the image share the same correlation with the output, a fact true for any given patch of size $s$ in the input. When a sufficient number of data is present, i.e. $P\geq P^*$, these correlations are larger than the sampling noise \textcolor{black}{ given by the finitess of the training set}. \textcolor{black}{These correlations can be used to group synonyms together.} Gradient descent \textcolor{black}{can capture these correlations}, constructing a representation invariant to synonyms exchange. \textcolor{black}{These correlations between synonyms and output are necessary for the network to solve the task: if a hierarchical model is designed without them, learning is essentially impossible (it requires an exponential number of data in the dimension)\cite{cagnetta2023deep}.} 
    
\end{itemize}

\begin{figure}[h]
    \centering
    \includegraphics[width=.5\textwidth]{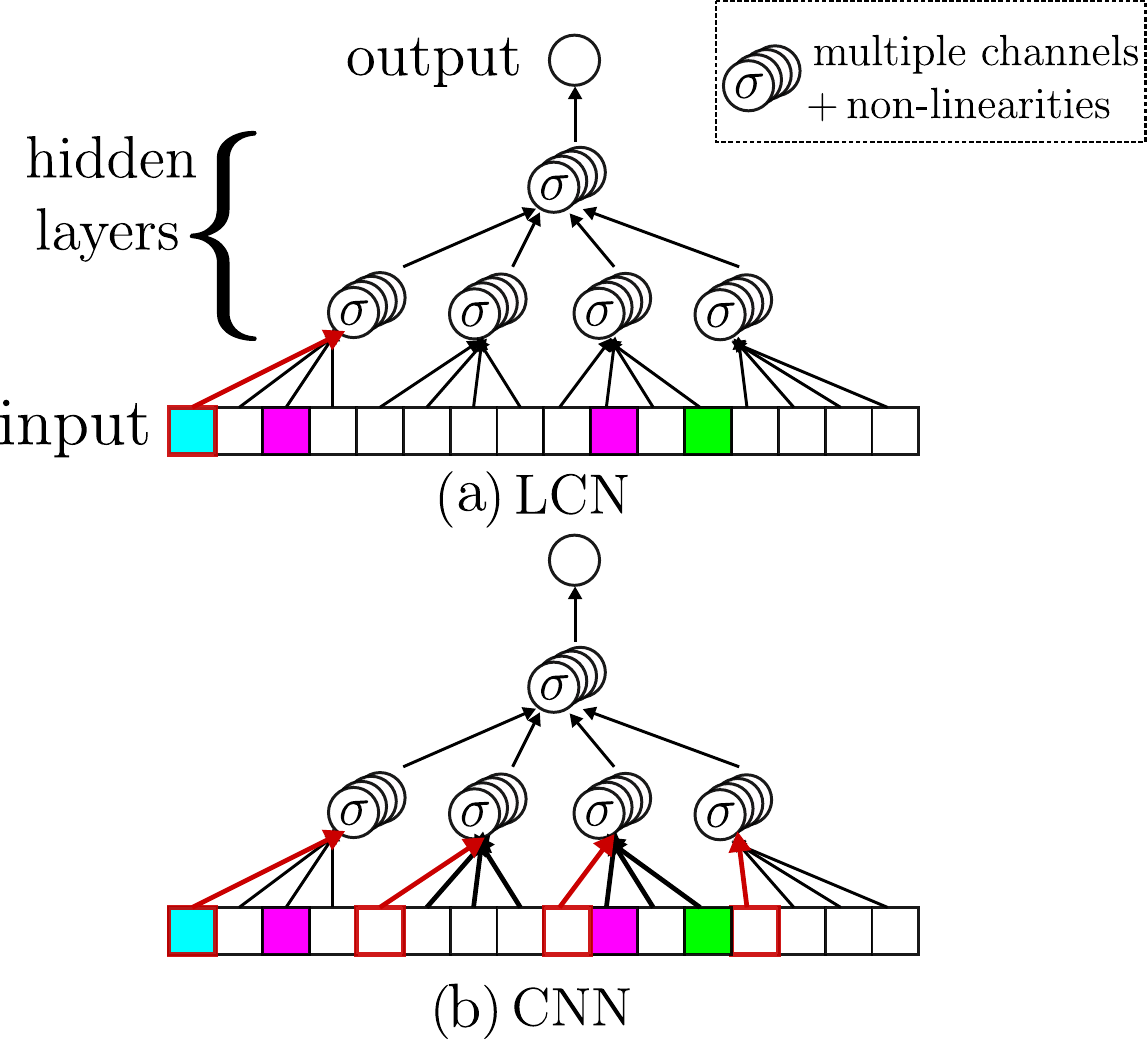}
    \caption{\textbf{Networks:} (a) Locally Connected Network (LCN). Each neuron's weight focuses on a single input element (in red). Networks have  $L$ hidden layers, with filters matching patches of size $s(s_0+1)$ from the generative process in \autoref{fig:rhm_all} (here $L=2$, $s=2$, $s_0=1$). A last fully connected layer connects the output of the last local layer with the output. (b) Convolutional Neural Network (CNN) with the structure of (a), featuring weight sharing such that each weight considers different pixels in all patches of size $s(s_0+1)$ (in red).}
    \label{fig:nets}
\end{figure}
\begin{figure*}[h]
    \centering
    \includegraphics[width=\textwidth]{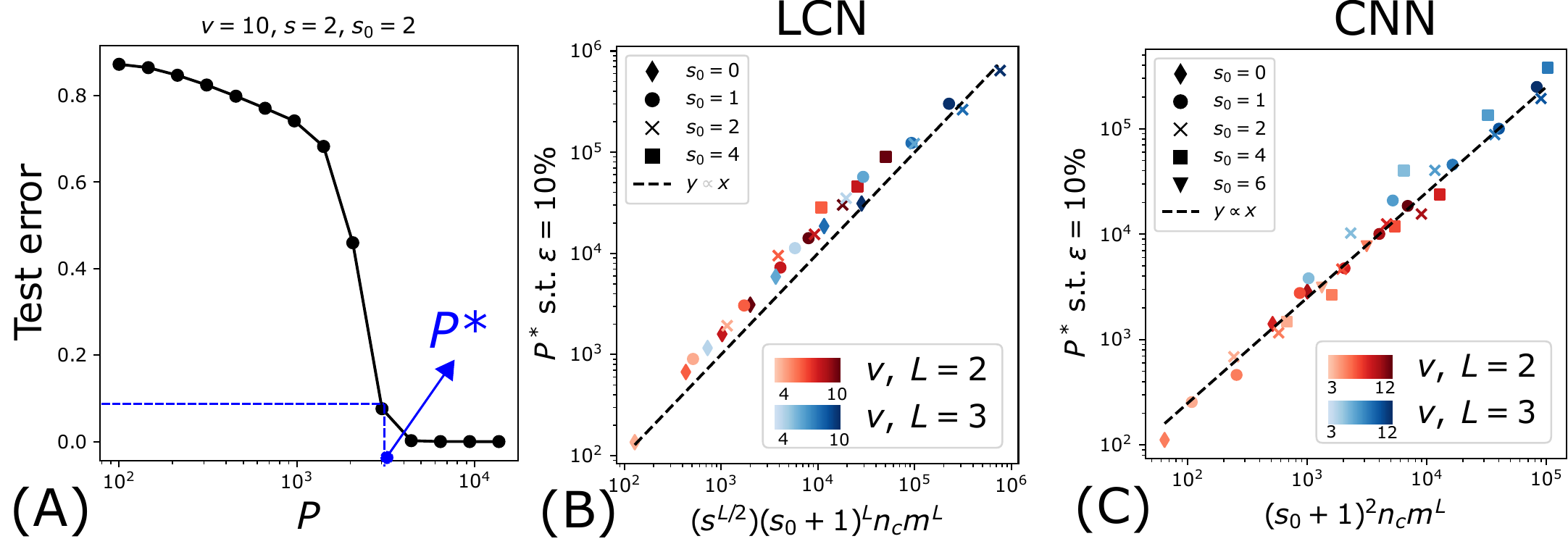}
    \caption{(A) Test error $\varepsilon(P)$  versus number of training points $P$. To extract the sample complexity $P^*$, we fix  an arbitrary threshold $\varepsilon^*=\varepsilon(P^*)$. Here $\varepsilon^*=10\%$. (B) Empirical sample complexity $P^*$ for a LCN to reach a $10\%$ test error $\varepsilon$ versus estimation of Eq. \ref{eq:pstar_LCN} for $s=2$, different depths $L$ (red for $L=2$, blue for $L=3$), different vocabulary sizes $v$ (different darkness), number of classes $n_c=v$, maximal $m=v^{s-1}$ and different $s_0$ (different markers). (C) Same as (B) for CNNs, supporting Eq. \ref{eq:pstar_CNN}. Further support for Eq. \ref{eq:pstar_LCN} and Eq. \ref{eq:pstar_CNN} is obtained by varying $s$, as shown in \autoref{app:sens_testerror_lcn}, \autoref{fig:tasksempos_lcn_s3} and \autoref{app:sens_testerror_cnn}, \autoref{fig:task_cnn_s3}.}
    \label{fig:procedure_sens}
\end{figure*}
\section{Sparsity and stability to diffeomorphisms} 
Our key insight is that spatial sparsity of features implies stability to diffeomorphisms. This point is illustrated using a sketched dog in \autoref{fig:rhm_all}, panel (2): if a few lines can define what is on a drawing, then small changes in the relative distances between these lines should not alter the class label. 


This idea can be readily implemented in generative models by adding an `uninformative'  feature 0 to each vocabulary $\mathcal{V}_{\ell}$ and imposing, for example, the constraint that each production rule in Eq. \ref{pro} and Eq. \ref{production} contains exactly $s\times s_0$ uninformative features. 
We implement sparsity in two ways, as shown in \autoref{fig:rhm_all}, panel (3). (A) Each of the $s$ informative features is embedded in a sub-patch of size $(s_0 + 1)$ with strictly $s_0$ empty elements. The position of each informative feature is independent of the other feature positions. (B) The $s$ informative features can occupy any 
position within the patch of $s(s_0 + 1)$ elements, as long as their order remains the same.  For both (A) and (B), at each level of the hierarchy, each uninformative feature will produce a patch of $s(s_0+1)$ uninformative features at the next level, as depicted in \autoref{fig:rhm_all}, panel (4). We denote model A as the Sparse Random Hierarchical Model (SRHM).

Note that (i) these processes generate very sparse data, with just $s^L$ informative features randomly placed in inputs of dimension $d=(s(s_0+1))^L$, as shown for a few examples in \autoref{fig:rhm_all}, panel (5). The informative features are represented with one-hot encoding with dimension $v$, while uninformative pixels are represented by empty columns, yielding input with size $d\times v$. (ii) Sparsity implies that some transformation leaves the task invariant. For (A), the transformations that do not affect the task include the motion of the $s$ informative features $\mu^1$ within patches of size $(s_0+1)$. For (B), any motion of the set of informative low-level features $\mu^1$ that leaves their order unchanged (as diffeomorphisms would do) does not alter the label, as illustrated in  \autoref{fig:rhm_all}, panel (3).  



\section{Sample complexity}
\label{sec:empirical}

\setcounter{footnote}{1}

We empirically analyze the number of training points required to learn the SRHM for both CNNs and for Locally Connected Networks (LCNs), a version of CNNs without weight sharing \cite{fukushima_cognitron_1975, Cun2012GeneralizationAN, favero_locality_2021}. 
In \cite{cagnetta2023deep}, it is shown that in the absence of sparsity, the sample complexity mildly depends on the architecture, as long as it is deep enough-- even for fully connected networks. Here, we start by restricting ourselves to architectures that match the generative process. Specifically, the LCN architecture has $L$ hidden layers with filter size and stride both equal to $s(s_0+1)$, \textcolor{black}{followed by a linear readout layer}, as shown in \autoref{fig:nets} (a).  The CNNs we consider are structured as the LCNs, but they implement weight sharing, as depicted in \autoref{fig:nets} (b). \textcolor{black}{By comparing the sample complexities of LCNs and CNNs we quantify the improvement achieved through the implementation of weight sharing}.  For details about the training scheme, we refer to \autoref{app:sens_testerror_lcn}. How common architectures such as VGG, ResNet, and EfficientNet learn the SRHM is investigated in \autoref{app:sens_testerror_newnets}. 

We utilize the SRHM to generate $P$ training points, corresponding to input-label pairs $(x,y)$ where $y=1,..,n_c$ and $x\in \mathbb{R}^{d\times v}$. An example of a learning curve representing the test error $\epsilon$ dependence on training set size $P$ is shown in \autoref{fig:procedure_sens} (A). The test error remains large with increasing $P$ until it decays rapidly to near-zero values at a certain scale $P^*$, corresponding to the sample complexity. To estimate it, we fix a threshold (e.g. $0.1$) and measure the minimal number of training points $P^*$ at which the test error achieves such a threshold. Modifying the threshold value does not qualitatively alter our results below. We focus on sparsity A, depicted in \autoref{fig:rhm_all} panel 3, as it is easier to analyze than the sparsity B. We show in \autoref{app:sparsity_B} that they lead to the same sample complexity.


We systematically apply this procedure to extract the sample complexity $P^*$ as the parameters $s$, $s_0$, $v$ and $L$ are changed while keeping the number of informative synonyms $m=v^{s-1}$ to its maximal value.

For the LCNs, the results shown in \autoref{fig:procedure_sens} (B) and \autoref{fig:tasksempos_lcn_s3} in \autoref{app:sens_testerror_lcn} indicate that: 
\begin{equation}
    P^*_{\text{LCN}}\sim C_{0}(s,L)(s_0+1)^L n_c m^L.
    \label{eq:pstar_LCN}
\end{equation}
\textcolor{black}{where our observations for $C_{0}(s,L)$ are consistent with $C_{0}(s,L)\sim s^{L/2}$.}
We will  motivate the dependence of $P^*_{\text{LCN}}$ with  $s_0$ in \autoref{sec:arguments}. 

For  CNNs, we observe in \autoref{fig:procedure_sens} (C) and \autoref{fig:task_cnn_s3} in \autoref{app:sens_testerror_cnn} that the sample complexity follows:
\begin{equation}
    P^*_{\text{CNN}} \sim C_{1}(s_0+1)^{2} n_c m^L.
    \label{eq:pstar_CNN}
\end{equation}
where $C_{1}$ is a constant.

\textcolor{black}{The relations} Eq. \ref{eq:pstar_LCN} and Eq. \ref{eq:pstar_CNN} are central to this work, as they specify how sparsity and combinatorial properties of the data affect the sample complexity. Remarkably, both sample complexities are exponential in $L$ and thus polynomial in the input dimension $d=(s(s_0+1))^L$, effectively beating the curse of dimensionality. Weight sharing proves to be beneficial for networks trained on the SRHM, since the sample complexity is only quadratic in $L$ in $(s_0+1)$ for CNNs, while it is exponential in $L$ for LCNs. We will argue why this is the case in \autoref{sec:arguments}.

\begin{figure}[h]
    \centering
    \includegraphics[width=.4\textwidth]{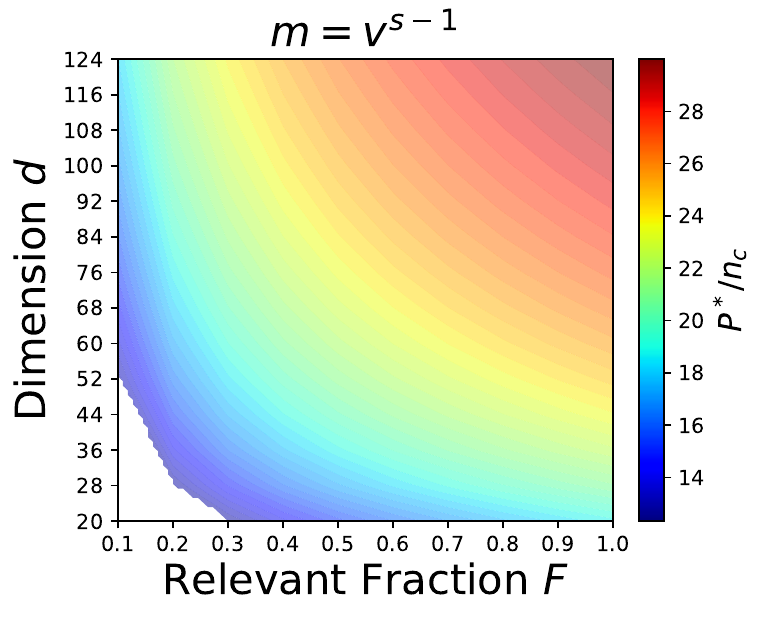}
    \caption{Sample complexity of LCN learning the SRHM for varying input dimension $d$ and input relevant fraction $F$ at the maximal case $m=v^{s-1}$, with $v=10$ and $s=5$, according to Eq. \ref{eq:pred_lcn_df}. The color map is in log scale. 
    At fixed dimension $d$, a smaller $F$ (hence higher sparsity) makes the task easier.}
    \label{fig:final}
\end{figure}

\textbf{Benefit of Sparsity.} We show that for locally connected nets, for reasonable assumptions and for a fixed dimension $d$, a higher sparsity is preferable as it leads to a smaller sample complexity. Introducing the fraction  of informative `pixels' in the image $F=(s_0+1)^{-L}$ as a measure of sparsity, we can reformulate the sample complexity $P^*_{LCN}$ in terms of the input dimension $d$ and image relevant fraction $F$, assuming that $m$ and $s$ are fixed but letting $L$ and $s_0$ change. One obtains:
\begin{equation}
    P^*_{\text{LCN}}\sim F^{\frac{\log m}{\log s}-\frac{1}{2}} d^{\frac{\log m}{\log s}+\frac{1}{2}},
    \label{eq:pred_lcn_df}
\end{equation}
which is indeed a growing function of $F$, as long as $m>\sqrt{s}$. This result is illustrated in \autoref{fig:final} for the case where $m$ takes its maximum value $m=v^{s-1}$. This indicates that neural networks can adapt to the sparsity of the task. 

 \begin{figure*}[h]
    \centering
    \includegraphics[width=.7\textwidth]{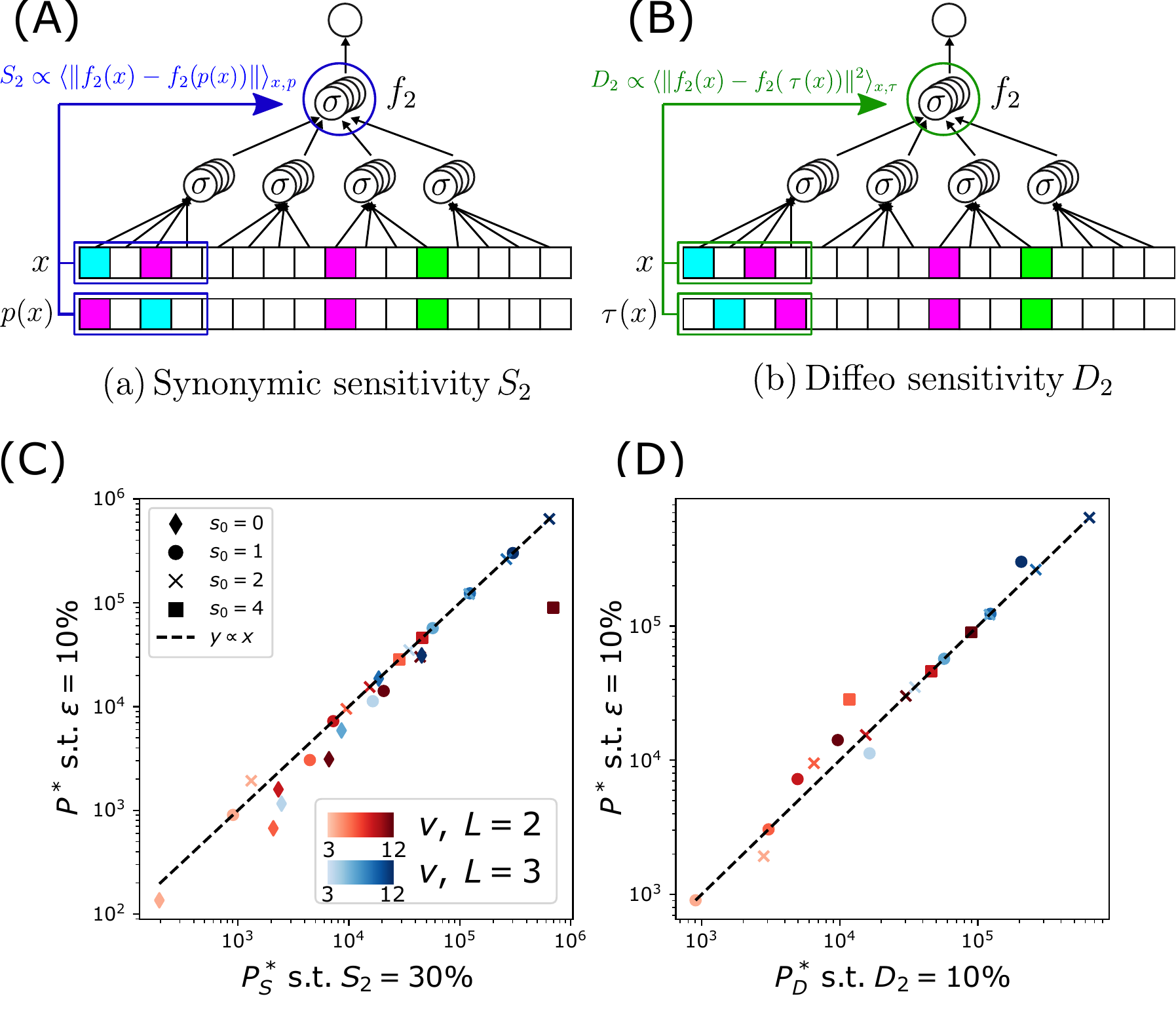}
    \caption{ \textbf{Top.} Procedure to compute the sensitivity $S_{2}$ and $D_{2}$, defined in Eq. \ref{eq:s_2} and Eq. \ref{eq:d_2}, illustrated for $L=2$. \textcolor{black}{We take a two-layer network, trained with $P$ training points on the sparse hierarchical dataset with $L=2$ levels. We apply either a (A) synonyms exchange $p$ or a (B) diffeomorphism $\tau$ on the input data $x$. Note that in (A) $p$ changes the features but not their position while in (B) $\tau$ changes the position of the features but not their value. Then we test the second net layer $f_2$ sensitivity to these transformations. }
    \textbf{Bottom.} (C) Empirical sample complexity $P^*$ to reach a $10\%$ test error $\varepsilon$ versus empirical sample complexity $P^*_S$ to reach $S_{2}=30\%$ for $s=2$, different depths $L$ (red for $L=2$, blue for $L=3$), different vocabulary sizes $v$ (different darkness), number of classes $n_c=v$, maximal $m=v^{s-1}$ and different $s_0$ (different markers). 
    (D)  Same as (C), for empirical sample complexity $P^*$ to reach a $10\%$ test error $\varepsilon$ versus empirical sample complexity $P^*_D$ to reach $D_{2}=10\%$. 
    The sensitivity thresholds have been tuned based on the form of $S_{2}$ and $D_{2}$ versus $P$, reported in \autoref{app:sens_testerror_lcn}, \autoref{fig:sens_lcn_L_2_s_2} for $L=2$. Both $P^*_S$ and $P^*_D$ are nearly equal to $P^*$.}
    \label{fig:tasksempos_lcn_s2}
\end{figure*}

\section{Learning invariant representation}
\label{sec:empirical_representation}

 In \autoref{sec:empirical}, we have shown that deep networks trained on the SRHM beat the curse of dimensionality, learning the task with a sample complexity polynomial in the input dimension. Here, we show that they manage to do so by learning representations that are insensitive to irrelevant aspects of the task. 

The first \textcolor{black}{invariance of the SRHM task} corresponds to the fact that different combinations of $s$ informative features are synonyms: to solve the task, it is not necessary to carry within the network the information of which actual synonym was present in the input. The second invariant considers transformations akin to diffeomorphisms, which shift the position of the informative features $\mu^{(1)}$. Formally, we measure such invariant by defining two operators acting on the input:

\begin{enumerate}
    \item Synonymic exchange operator $p$. This operator takes in a datum $x$ and substitutes each informative $s-$patch of features $\mu^{(1)}$, produced by a given $\mu^{(2)}$, with one of its $m-1$ synonyms, chosen uniformly at random, keeping the feature positions intact.  \autoref{fig:tasksempos_lcn_s2} (A) illustrates the action of $p$. The sensitivity of the representation at a given hidden layer $f_{k}$ to the action of $p$ is measured by the quantity $S_{k}$:
    \begin{equation}
        S_{k}=\frac{\langle||f_{k}(x)-f_{k}(p(x))||^2\rangle_{x,p}}{\langle||f_{k}(x_1)-f_{k}(x_2))||^2\rangle_{x_1,x_2}},
        \label{eq:s_2}
    \end{equation}
    where $x$, $x_1$, and $x_2$ belong to a test set, and the average $\langle .\rangle$ is over the test set and on the random exchange of synonyms $p$. A visualization of $S_{k}$ for $k=2$ for a two-layer network is shown in \autoref{fig:tasksempos_lcn_s2} (A). If the internal representation $f_{k}$ has learnt the synonyms associated with production rule Eq. \ref{production} at level $\ell =2$, then $S_k$ is small. \textcolor{black}{The sensitivity of the network output to $p$, shown in \autoref{fig:phasediagram_new} (D, F), is defined similarly to \eqref{eq:s_2}.}
    
    \item  A discretised diffeomorphism operator $\tau$. This operator takes as input a datum $x$ and randomly shifts its informative features according to the possible positions obtained when the features $\mu^{(2)}$ generate patches of features $\mu^1$, as illustrated in \autoref{fig:rhm_all}, panel (3). In this process, the nature of the feature remains the same. In \autoref{fig:tasksempos_lcn_s2} (B) the action of $\tau$ on a datum $x$ is shown. The sensitivity $D_{k}$ of $f_k$ to diffeomorphisms $\tau$ is defined as:
    \begin{equation}
        D_{k}=\frac{\langle||f_k(x)-f_k(\tau(x))||^2\rangle_{x,\tau}}{\langle||f_k(x_1)-f_k(x_2))||^2\rangle_{x_1,x_2}}.
        \label{eq:d_2}
    \end{equation}
    If $f_k$ has learnt the invariance to diffeomorphisms, then $D_k$ is zero. This definition of sensitivity is akin to what is used for images \cite{petrini_relative_2021}. \textcolor{black}{The sensitivity of the network output $f$ to $\tau$, shown in \autoref{fig:phasediagram_new} (C, E), is defined analogously to \eqref{eq:d_2}.}
\end{enumerate}

It is possible to generalize the sensitivities Eq. \ref{eq:s_2} and Eq. \ref{eq:d_2} to measure how much the representation at any hidden layer $k$ of a network\textcolor{black}{, or its output,} is sensitive to exchange of synonyms or diffeomorphisms related to the \textcolor{black}{features $\mu^{(\ell)}$ produced} by the latent $\mu^{(\ell+1)}$.  We show in \autoref{app:sens_testerror_lcn} and \autoref{app:sens_testerror_cnn} that the representation at layer $k\geq 2$ becomes insensitive to transformations applied to levels $\ell \leq k-1$ of the hierarchy, for training set size $P\gg P^*$. \textcolor{black}{At that point, the output becomes insensitive to transformations applied at any level $\ell$ of the hierarchy.}


Here, we focus on $S_{2}$ and $D_{2}$ (corresponding to the case $k=2$ and $\ell=1$). We measure how these quantities depend on the size of the training set, as shown in \autoref{app:sens_testerror_lcn}, \autoref{fig:sens_lcn_L_2_s_2}, and in \autoref{app:sens_testerror_cnn}, \autoref{fig:sens_cnn_L_2_s_2}, for $L=2$. Subsequently, we define sample complexities $P^*_S$ or $P^*_D$ associated with $S_{2}$ and $D_{2}$, using a similar procedure as for the test error: we measure the training set size where these quantities reach some threshold value. 

\autoref{fig:tasksempos_lcn_s2} (C) and (D) present our key result for LCNs:
\begin{equation}
    P_S^*\approx P^*_{\text{LCN}}, \quad P_D^*\approx P^*_{\text{LCN}},
    \label{eq:sempos_scale}
\end{equation}
further supported by \autoref{fig:sem_pos_lcn} in \autoref{app:sens_testerror_lcn} for a different value of $s$. In essence, the insensitivities to diffeomorphisms and synonyms exchange are acquired for the same training set size, precisely when the network learns the task. This observation appears to be universal, as it holds for common convolutional architectures like VGG, ResNet, and EfficientNet as demonstrated in \autoref{app:sens_testerror_newnets}, \autoref{fig:sens_newnets}, for the simple convolutional architectures pictured in \autoref{fig:nets} (b) in \autoref{app:sens_testerror_cnn}, \autoref{fig:sem_pos_cnn} \textcolor{black}{ and even for fully-connected networks in \autoref{app:fcn}, \autoref{fig:fcn}.}

Therefore, {\it deep networks learn to be insensitive to diffeomorphisms as they learn a hierarchical representation}, \textcolor{black}{which is thought to be crucial to achieve high performance}. This central result rationalizes the observed correlation between sensitivity to diffeomorphisms \textcolor{black}{of the network output} (\textcolor{black}{thus indicative of building a hierarchical representation of the data,) } and test error displayed in \autoref{fig:phasediagram_new} (A) for CIFAR10 and (C) in the SRHM. 

\textcolor{black}{The relationship between insensitivity to synonyms and to diffeomorphisms also appears when considering how these quantities evolve with the size of the training set $P$,} as documented in \autoref{fig:phasediagram_new} (B, E, F). \textcolor{black}{The results shown in \autoref{fig:phasediagram_new} (C,D,E,F) for the network output hold also for the sensitivities of internal layers, as shown in \autoref{fig:phasediagram_new_int} in \autoref{app:corr_int}.}

\section{Sample complexities arguments}
\label{sec:arguments}




In the absence of sparsity $s_0=0$, the sample complexity essentially corresponds to the training set size necessary for detecting synonyms \cite{cagnetta2023deep}. \textcolor{black}{Following this result,} we present a simple heuristic argument for the sample complexity of the LCN architecture in the sparse case $s_0>0$. \textcolor{black}{Crucially, this argument also explains why invariance to synonyms and to smooth transformations are learnt together.}
\begin{itemize}
    \item  \textcolor{black}{Any given 2-level latent variable $\mu^{(2)}$ closest to the input can generate different synonyms, distorted with different diffeomorphisms due to sparsity, as in \autoref{fig:rhm_all}, panel (3).}
    
    \item \textcolor{black}{All these different representations produced by that latent variable $\mu^{(2)}$ have the same correlation with the class label.}
    
    \item \textcolor{black}{This correlation can be used to group together the  representations produced by $\mu^{(2)}$, if the training set is larger than some sample complexity.  We can estimate this sample complexity for LCNs as follows. A single weight connected to the input will not see any information for most data, as it will most often see the "pixel" or low-level feature $\mu^{(1)}=0$, which is uninformative. With respect to the case $s_0=0$,  the fraction of data with local information is diminished by a factor $(s_0+1)^{-L}$. To detect correlations beyond sampling noise, the sample complexity has thus to be increased by a factor $(s_0+1)^{L}$, as empirically observed in Eq. \ref{eq:pstar_LCN}. We show in \autoref{app:GD} that at this sample complexity a single step of GD can aggregate the equivalent representations produced by $\mu^{(2)}$. }
    
    \item \textcolor{black}{The hidden representation in the first hidden layer  thus becomes insensitive to diffeomorphisms and synonyms at the same sample complexity, as shown in \autoref{fig:tasksempos_lcn_s2} (C, D). Once the representations that encode for features $\mu^{(2)}$ have been detected, the problem is much simpler, corresponding to a generative model with $L-1$ levels instead of $L$. Thus, representations of higher production rules $\mu^\ell$, $\ell>2$ are also detected and represented together in the higher layers of the network beyond that characteristic training set size. Indeed, synonyms and insensitivity to diffeomorphisms are learnt at all levels of the hierarchy in one go, as illustrated in \autoref{app:sens_testerror_lcn}, \autoref{fig:sens_lcn_L_2_s_2}. }    
\end{itemize}



Qualitatively, the same scenario holds for CNNs. One expects a different sample complexity since each weight is now connected to a fraction of the input that is independent of $L$. Yet, the quadratic dependence on $s_0+1$ remains to be understood.

\section{Limitations}

\textcolor{black}{We predicted that good performance, insensitivity to diffeomorphism, and insensitivity to synonyms exchange should occur concomitantly as the training set size is increased. However to test this prediction on benchmark image datasets, we are currently missing an empirical procedure to measure insensitivity to synonyms. To perform such a measurement, one needs to be able to change what composes a given image at various scales. The recent result of \cite{sclocchi2024phase} suggest that change of low-level features can be obtained using diffusion-based generative models, and that the scale where the composition of the image changes is controlled by the magnitude of the noise that enters in these methods. It would be interesting to use this result to test our prediction. Another approach would be to test stability to actual synonyms in text data, and quantify how this property correlates with the performance of natural language processing systems. \looseness-1}

\section{Conclusions}

\textcolor{black}{Understanding} the nature of real data is an elusive problem yet central to many quantitative questions of the field.
We have unified in a common framework two  \textcolor{black}{different} approaches to this problem: the first assumes the data to be combinatorial and hierarchical, while the second emphasizes the task insensitivity with respect to smooth transformations,  relevant for example for image datasets. This framework was obtained by introducing models that display both properties, based on the notion that sparsity of informative features in space naturally endows stability to smooth transformations. These models explain the strong empirical correlation between stability to smooth transformations and performance, and further predict that both quantities should correlate with the insensitivity to discrete changes of low-level features in the data. 


Finally, although we focused on classification problems, the generative models we introduced display a rich structure in the data distribution $P(x)$ itself, which is not only hierarchical but \textcolor{black}{is invariant } to smooth transformations. These models are thus promising to understand how unsupervised methods beat the curse of dimensionality, by composing or deforming features they have learnt from examples.

\section*{Impact statement}

This paper presents work whose goal is to advance the field of Machine Learning. There are many potential societal consequences of our work, none of which we feel must be specifically highlighted here.

\section*{Acknowledgements}

The authors thank Francesco Cagnetta, Alessandro Favero, Leonardo Petrini and Antonio Sclocchi for fruitful discussions and helpful feedback on the manuscript. This work was supported by a grant from the Simons Foundation (\#454953
Matthieu Wyart).

\bibliography{main}
\bibliographystyle{icml2024}

\newpage
\onecolumn
\appendix
\begin{center}
{\LARGE \textbf{Appendix}}
\end{center}
\section{Sparsity B}
\label{app:sparsity_B}
\begin{figure}[h]
    \centering
    \includegraphics[width=.67\textwidth]{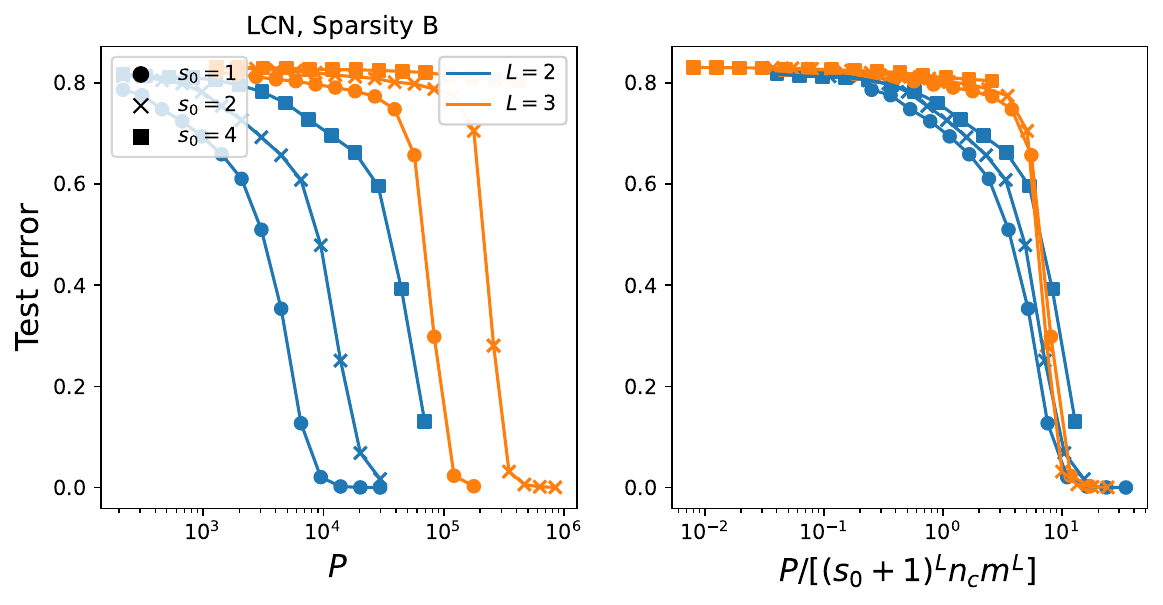}
    \caption{Left: test error versus number of training points $P$ of LCN trained on the SRHM with sparsity B (described in \autoref{fig:rhm_all}, panel (3)), with different $L$ (different colors), different $s_0$ (different  markers), $s=2$ and $m=v=n_c=6$. Right: same as left but rescaling $P$ by Eq. \ref{eq:pstar_LCN}.}
    \label{fig:sparsB_lcn}
\end{figure}

\begin{figure}[h]
    \centering
    \includegraphics[width=.67\textwidth]{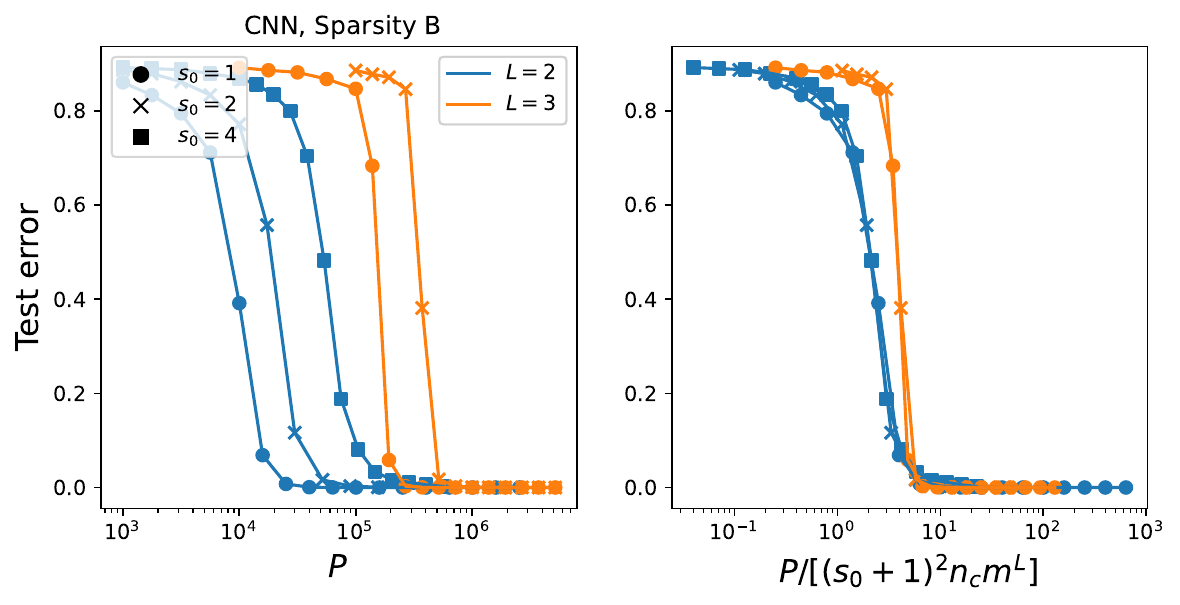}
    \caption{Left: test error versus number of training points $P$ of CNN trained on the SRHM with sparsity B (described in \autoref{fig:rhm_all}, panel (3)), with different $L$ (different colors), different $s_0$ (different  markers), $s=2$ and $m=v=n_c=6$. Right: same as left but rescaling $P$ by prediction Eq. \ref{eq:pstar_CNN}.}
    \label{fig:sparsB_cnn}
\end{figure}

In this Appendix, we show support for the robustness of the sample complexities Eq. \ref{eq:pstar_LCN} and Eq. \ref{eq:pstar_CNN} to the choice of the sparsity in \autoref{fig:rhm_all}, panel (3). Indeed, we observe in \autoref{fig:sparsB_lcn} and \autoref{fig:sparsB_cnn} that the learning curves obtained implementing Sparsity B in the data are consistent with Eq. \ref{eq:pstar_LCN} and Eq. \ref{eq:pstar_CNN}, obtained from data generated with Sparsity A.


\section{Common architectures learning the SRHM}
\label{app:sens_testerror_newnets}

\textbf{Architectures.} All networks implementations can be found at \href{https://github.com/leonardopetrini/diffeo-sota/tree/main/models}{github.com/leonardopetrini/diffeo-sota/tree/main/models}. In Table \ref{tab:nets}, adapted with permission of the authors from \cite{petrini_relative_2021}, we list the structural choices of those networks.

\textbf{Training scheme.} We use Stochastic Gradient Descent (SGD) as optimizer, with batch size 4 and momentum 0.9. The learning rate $lr$ has been optimized by extensive grid search, finding: $lr = 10^{-4}$. We use as train loss the cross entropy loss, and we stop the training when it reaches a threshold of $10^{-2}$. 

\begin{figure}[h]
    \centering
    \includegraphics[width=1\textwidth]{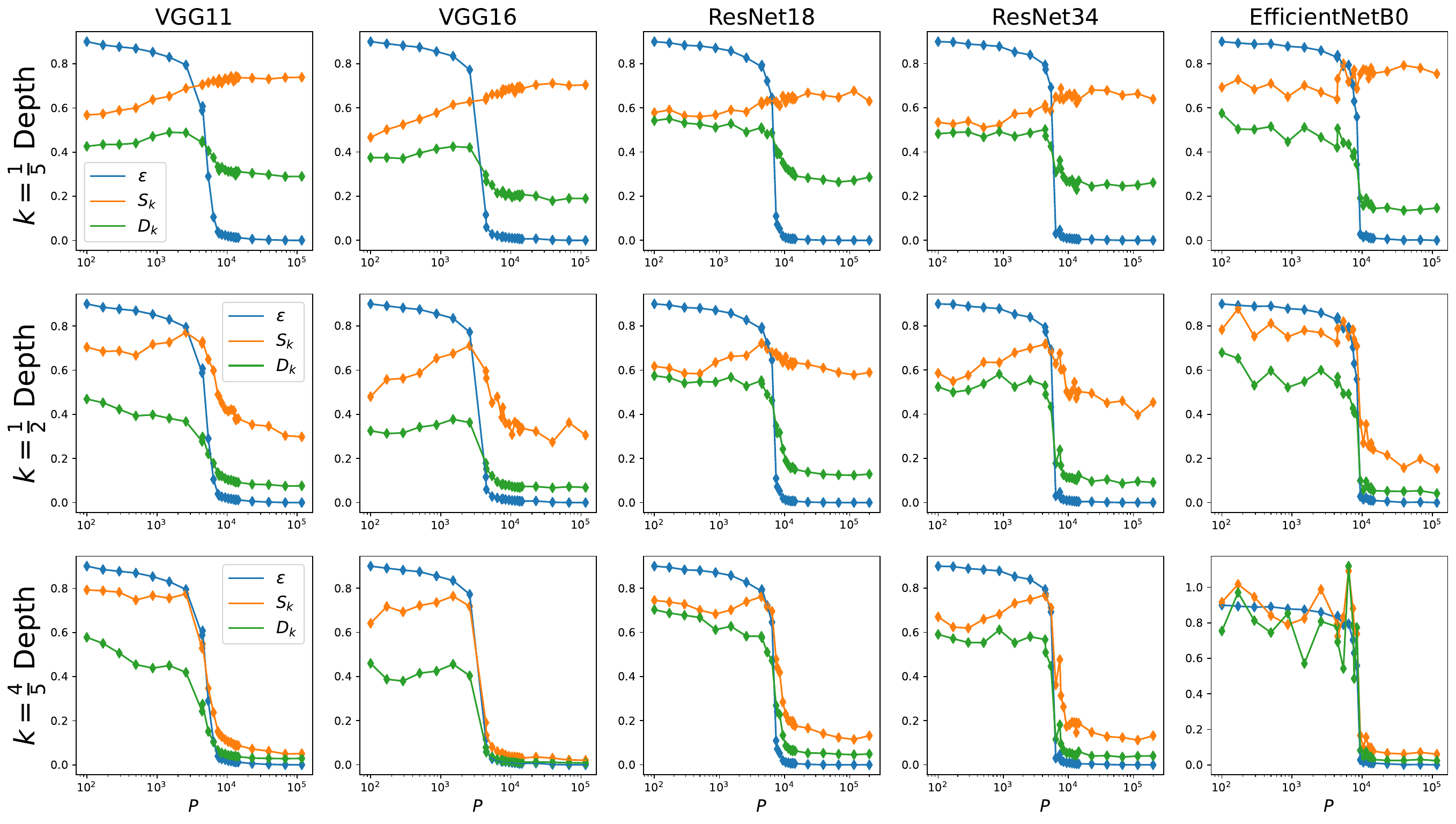}
    \caption{Test error (in blue), sensitivity to synonymic exchange (in orange), sensitivity to diffeomorphisms (in green) for increasing number of training points $P$ of different common architectures  (different columns). The sensitivities compute how much internal representations, at increasing relative depth $k$ for increasing row, are sensitive to transformations applied at the input. Note that at $80\%$ relative depth all the networks reach small sensitivities and test error for large enough $P$. We choose this relative depth for the results shown in \autoref{fig:phasediagram_new} (A,B). The sensitivities $S_{k}$ and $D_{k}$ refer to the case $\ell=1$ in Eq. \ref{eq:s_kl} and Eq. \ref{eq:d_kl}.}    
    \label{fig:sens_newnets}
\end{figure}

\textbf{Figures.} In \autoref{fig:sens_newnets} we report the test error and the sensitivities to input transformations of 5 different common convolutional networks with respect to the number of training points $P$. The sensitivities are computed for 3 different layers at different relative depths.

\begin{table}[h!]
  \begin{center}
  \caption{\textbf{Network architectures, main characteristics.} For each column we list the salient structures of a different architectures used in \autoref{fig:sens_newnets}. Table adapted with permission of the authors from \cite{petrini_relative_2021}\\}
  \label{tab:nets}
  \small{
    \begin{tabular}{c c c c}
        \toprule
        &\\
      \textit{structures} & \textbf{VGG} & \textbf{ResNet} & \textbf{EfficientNetB0-2}\\
      & \tiny{\cite{simonyan_very_2015}} & \tiny{\cite{he_deep_2016}} & \tiny{\cite{tan_efficientnet_2019}} \\
      \midrule
      depth  & 11, 16 & 18, 34 & 18 \\
      num. parameters  & 9-15 M & 11-21 M & 5 M \\
      FC layers &  1 & 1 & 1 \\
      activation &  ReLU & ReLU & swish \\
      pooling & max & avg. (last layer only) & avg. (last layer only) \\
      dropout & / & / & yes + dropconnect \\
      batch norm & if 'bn' in name & yes & yes \\
      skip connections & / & yes & yes (inv. residuals) \\
      \bottomrule
    \end{tabular}
    }
  \end{center}
\end{table}

\section{Learning the SRHM with Gradient Descent}
\label{app:GD}

In hierarchical generative models, it is known that the first step of gradient descent in some simplified setup can already be sufficient to group together lowest-level synonyms   \cite{malach2020implications,cagnetta2023deep}. In particular, the case of the SRHM with $s_0=0$ has been analyzed in \cite{cagnetta2023deep}. Here we show for LCNs with a given initialization that the statistics of the first step of gradient descent for $s_0>0$ is equivalent to the case $s_0=0$, with a training set size reduced by a factor $(s_0+1)^{L}$, thus supporting the result of Eq. \ref{eq:pstar_LCN} on sample complexity.


We consider an instance of the SRHM with $L$ levels, from which we generate $P$ training points ${x_k}$ with label $y(x_k)$. We recall that training points have dimension $d=(s(s_0+1))^L$, with each one of the  informative $s^L$ features being represented with one-hot encoding of $v$ features, while the $(s_0+1)^L$ uninformative features by empty columns of dimension $v$. The labels are represented with a one-hot encoding of the $n_c$ labels.  To learn the SRHM, we use as a network a LCN with $L$ hidden layers, followed by a linear layer. Each one of the hidden layers $f_k$ for $k\in\{1,...,L\}$ is defined with respect to the previous layer $f_{k-1}$. We use filters with size and stride equal to $s(s_0+1)$, yielding a reduction of $s(s_0+1)$ in the spatial size of the hidden layer at each layer. Each entry of $f_k$ is specified by two indices, one for the channel $c\in\{1,...,H_{k}\}$ and one for the spatial location $i\in\{1,...,(s(s_0+1))^{L-k}$, and it is given by:
\begin{equation} 
   \left[f_{k}(x)\right]_{c; i} = \phi\left(\frac{1}{\sqrt{H_{k-1}}}\displaystyle\sum_{c'=1}^{H_{k-1}} \vec{w}_{c,c',i}^k \cdot \vec{p}_{i}\left(\left[f_{k-1}(x)\right]_{c'}\right)\right), 
    \label{eq:convolutional_layer}
\end{equation}
where $\phi$ is the ReLU non-linearity function and $\vec{w}_{c,c',i}^k$ are the filters of the $c-$th channel of the $k$-th layer. Each filter $\vec{w}_{c,c',i}^k$ connects channel $c$ of layer $k$ with the patch $\vec{p}_i$ channel $c'$ of layer $k-1$. Note that the filters depend on the patch location $i$, differently with respect to CNNs (see \autoref{fig:nets}). The vector $\vec{p}_{i,j}\left(\left[f_{k-1}(x)\right]_{c'}\right)$ denotes a $s(s_0+1)$-dimensional patch of $\left[f_{k-1}(x)\right]_{c'}$ centered in the spatial location $i$ at channel $c'\in\{1,..,H_{k-1}\}$.  The bottom layer $k=1$ is directly looking at the input, hence we define $f_0(x)$ as the identity and the number of channels $H_1$ is equal to $v$. The output of the LCN is given by:
\begin{equation}
    f(x)_\alpha = \frac{1}{H_L}\sum_{c=1}^{H_L} a_{\alpha,c} [f_L(x)]_c,
\end{equation}
with $f(x)$ being a vector of dimension equal to the number of classes $n_c$ and $\{a_{\alpha,c}\}$ the parameters of the last linear layer. We train such a LCN with GD on the cross-entropy loss
\begin{equation}
\label{eq:cross-ent-loss}
\mathcal{L} = \sum_{k=1}^P \left[ -\displaystyle\sum_{\beta=1}^{n_c} y(x_l)_{\beta} \log{\left(\frac{e^{\left(f(x_k)\right)_{\beta}}}{\sum_{\beta'=1}^{n_c} e^{\left(f(x_k)\right)_{\beta'}}}\right)} \right].
\end{equation}
For simplicity we (i) take all the number of channels $H_k$ equal to $H$ for $k>1$, (ii) send $H\rightarrow\infty$, (iii) fix the linear layer parameters $a_{\alpha,c}$ to be i.i.d. standard Gaussian variables and (iv) initialize all the weights $\vec{\omega}_{c,c',i}^k$ to be equal to the unitary vector $\vec{1}$ renormalized by $\sqrt{H}$. As a consequence of (iii), the network output $f(x)_\alpha$ is zero identically at initialization. 

We now derive the weight updates after one step of GD. Let's focus on the weights at the bottom layer $[\omega^1_{c,c,',i}]_{i_0}$, where $i_0$ denotes a position within the filter of size $s(s_0+1)$ and $i\in\{1,...,[s(s_0+1)]^{L-1}\}$. Note that, since there is no weight sharing, each position $z$ within the input dimension $d$ is seen just one weight at the bottom layer, with $z=z(i,i_0)=(i-1)(s(s_0+1))+i_0$. Consequently, the weight update for a weight $[\omega^1_{c,c,',i}]_{i_0}$ just depends on the content of such pixel location $z=z(i,i_0)$:
\begin{equation}
\begin{aligned}
    \frac{ \partial \Delta[\omega^1_{c,c,',i}]_{i_0}}{\partial t}&= -\nabla_{[\omega^1_{c,c,',i}]_{i_0}}\mathcal{L}\\ 
    \,&=-\frac{1}{P}\sum_{k=1}^P\left[\sum_{\alpha=1}^{n_c}\left(\frac{1}{n_c}-(y(x_k))_\alpha\right)\frac{1}{H}\sum_{h_L=1}^H a_{\alpha,h_L}\frac{1}{H}\sum_{h_{L-1}=1}^H ... \frac{1}{H}\sum_{h_{2}=1}^H[x_k]_{c',z(i,i_0)}\right]\\
    \,&=-\frac{1}{P}\sum_{k=1}^P\left[\sum_{\alpha=1}^{n_c}\left(\frac{1}{n_c}-(y(x_k))_\alpha\right)\frac{1}{H}\sum_{h_L=1}^H a_{\alpha,h_L}\right][x_k]_{c',z(i,i_0)}.
    \label{update}
\end{aligned}
\end{equation}
On average over all the training sets, the pixel at location $z(i,i_0)$ of the input datum $x_k$ relates to an informative feature with probability $p_L=(s_0+1)^{-L}$. Consequently, on average just a fraction $P'=p_L P$ of training points will give a non-vanishing contribute to the weight update Eq. \ref{update}, which will be identical to the GD equation of an instance of the sparseless SRHM with $P'$ training points. Thus, the sample complexity in the sparse case $s_0>0$ is equal to the sparseless sample complexity $n_c m^L$ increased by a factor $(s_0+1)^{L}$, as empirically observed in Eq. \ref{eq:pstar_LCN}. At this sample complexity, a single step of GD is capable to group together representations with equal correlation with the label, as shown in \cite{cagnetta2023deep}. \looseness -1

\section{Sample complexities and learnt representations for LCNs}
\label{app:sens_testerror_lcn}
\textbf{Architectures.} Shown in \autoref{fig:nets} (a). All layers consist of 512 channels. 

\textbf{Training scheme.} We use Stochastic Gradient Descent (SGD) as optimizer, with batch size 4 and momentum 0.9. The learning rate $lr$ has been optimized by extensive grid search, finding: $lr = 0.01$ for $s=2$ and $s_0<4$, $lr = 0.01$ for $s=2$ and $s_0\geq 4$ and $lr = 0.003$ for $s\geq 2$. We use as train loss the cross entropy loss, and we stop the training when it reaches a threshold of $10^{-3}$. 

\begin{figure}[h]
    \centering
    \includegraphics[width=.67\textwidth]{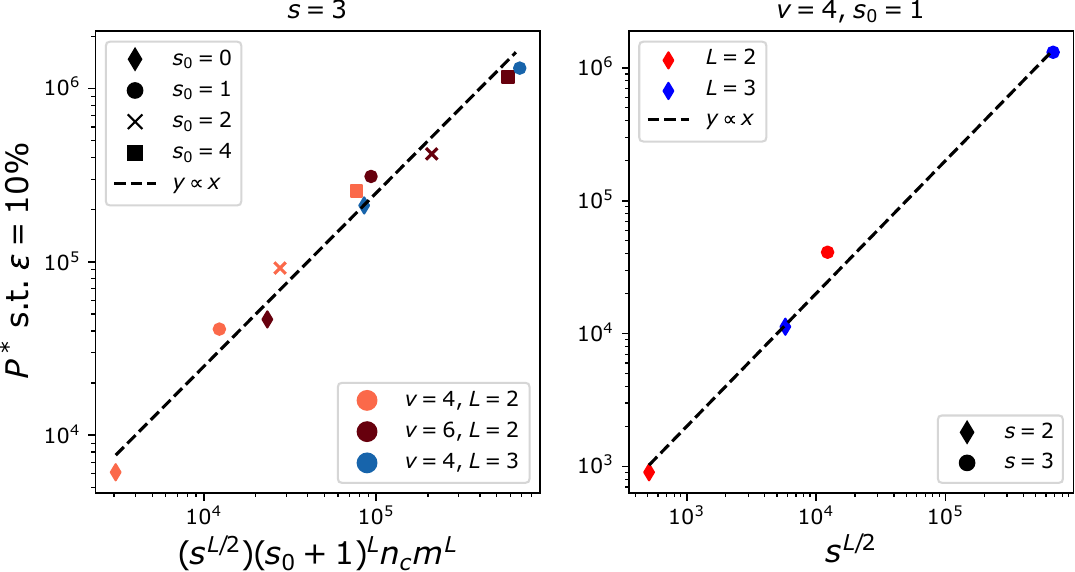}
    \caption{\textbf{LCN:} Left panel: empirical sample complexity $P^*$ to reach a $10\%$ test error $\varepsilon$ versus prediction Eq. \ref{eq:pstar_LCN} for $s=3$, different vocabulary sizes $v$ and different depths $L$ (different colors), number of classes $n_c=v$, maximal $m=v^{s-1}$ and different $s_0$ (different markers).Note an additional factor $s^{L/2}$ in the prediction, further supported by the right panel, realised at fixed $s_0=1$, $n_c=m=4$, and varying $s$ and $L$. }
    \label{fig:tasksempos_lcn_s3}
\end{figure}

\begin{figure}[h]
    \centering
    \includegraphics[width=.7\textwidth]{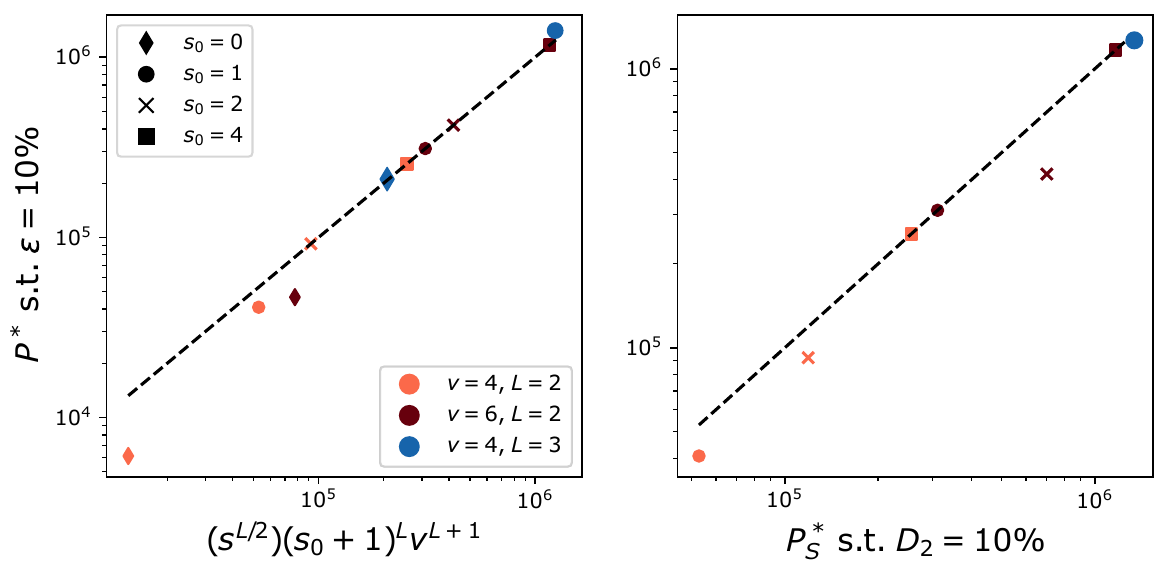}
   \caption{\textbf{LCN.} Left panel: empirical sample complexity $P^*$ to reach a $10\%$ test error $\varepsilon$ versus empirical sample complexity $P^*_S$ to reach $S_{2}=50\%$ (as defined in Eq. \ref{eq:s_2}) for $s=3$, different vocabulary sizes $v$ and different depths $L$ (different colors), number of classes $n_c=v$, maximal $m=v^{s-1}$ and different $s_0$ (different markers). Right panel: same as left for the sample complexity $P^*_D$ to reach a $D_{2}=10\%$ (with $D_{2}$ as defined in Eq. \ref{eq:d_2}). The thresholds have been tuned based on the form of $S_{2}$ and $D_{2}$ versus $P$. }    
    \label{fig:sem_pos_lcn}
\end{figure}

\textbf{Sample complexity figures.} In \autoref{fig:tasksempos_lcn_s3}, realised with $s=3$, we show additional support to the scaling of the sample complexity $P^*_{LCN}$, defined in Eq. \ref{eq:pstar_LCN}. In \autoref{fig:sem_pos_lcn} we show for $s=3$ that the insensitivity to diffeomorphism and to synonyms exchange are achieved at the same sample complexity $P^*_{LCN}$ at which the task is learned, supporting Eq. \ref{eq:sempos_scale}.

\textbf{Learnt representations figures.} To assess whether the internal representation of the trained network have learnt the production rules Eq. \ref{pro} and Eq. \ref{production}, we define an operator $p_l$ which takes in a datum $x$ and substitutes each informative $s$ patches at generation level $l$ with one of its $m-1$ synonyms, chosen uniformly at random, keeping the feature positions intact. The level $l$ goes from $l=1$, the input, to $L$, the patch closest to the label. We define the sensitivity $S_{k,l}$ of the network representation $f_k$ at an internal layer $k\in\{1,...,L\}$ (with $k=1$ being the layer closest to the input while $k=L$ the one closest to the output) to synonymic exchange $p_l$ as its average change before and after applying $p_l$ on an input datum $x$:
\begin{equation}
    S_{k,l}=\frac{\langle||f_k(x)-f_k(p_l(x))||^2\rangle_{x,p_l}}{\langle||f_k(x_1)-f_k(x_2))||^2\rangle_{x_1,x_2}},
    \label{eq:s_kl}
\end{equation}
where $x$, $x_1$ and $x_2$ belong to a test set, while the average $\langle .\rangle$ is over the test set and on the random exchange of synonyms $P_l$. The same definition apply for the output of the deep network. The measure Eq. \ref{eq:s_2} is a particular case of Eq. \ref{eq:s_kl} for $k=2$ and $l=1$, which can be visualized in \autoref{fig:tasksempos_lcn_s2} (A). 

\begin{figure}[h]
    \centering
    \includegraphics[width=\textwidth]{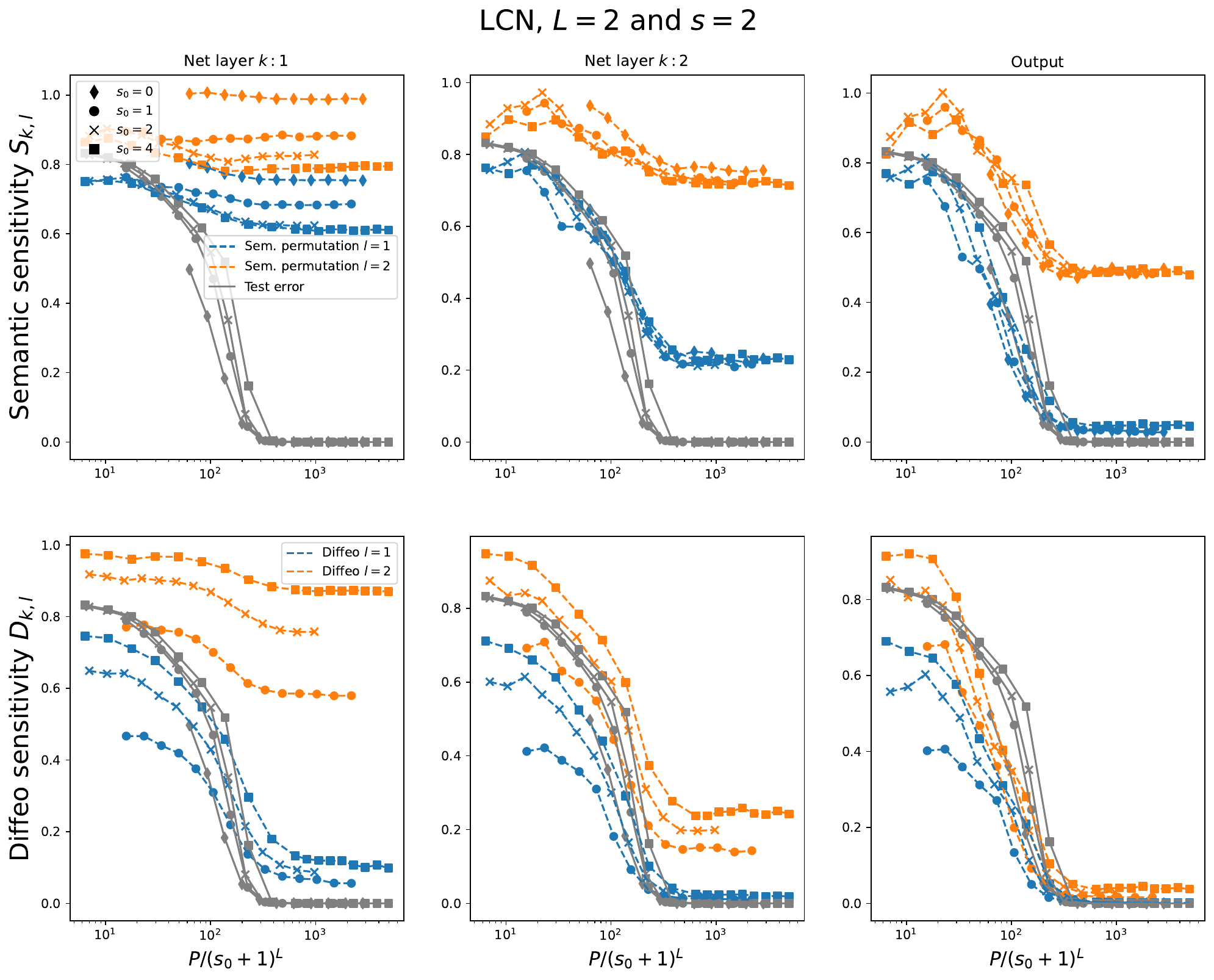}
    \caption{Top: sensitivity to synonyms exchange $S_{k,l}$ Eq. \ref{eq:s_kl} of a $L-$layer LCN, trained on the SRHM with $L=2$, $s=2$ and $m=v=n_c=8$, versus number of training points $P$ rescaled by prediction Eq. \ref{eq:pstar_LCN}. Going from left to right column the network layer $k$ increases from the layer closest to the input to the output. For each column at fixed $k$ there are plotted in color the sensitivities $S_{k,l}$ to synonyms exchange at the data level $l\in[1,...,L]$, and in grey the test error. Different markers stand for different $s_0$. Bottom: same as top, but with the sensitivity to diffeomorphisms $D_{k,l}$ Eq. \ref{eq:d_kl}. The sensitivities $S_2$ and $D_2$ defined in Eq. \ref{eq:s_2} and Eq. \ref{eq:d_2} relate to the case $k=2$ and $l=1$ here.
    }
    \label{fig:sens_lcn_L_2_s_2}
\end{figure}

To estimate the sensitivity to diffeomorphisms of trained networks we define an operator $\tau_l$, which takes as input a datum $x$ and shifts randomly its features according to the possible positions shown in \autoref{fig:rhm_all}, panel (3), not impacting the feature values. Similarly to $S_{k,l}$, we define the sensitivity of $f_k$ to diffeomorphisms $\tau_l$ at layer $l$ as
\begin{equation}
    D_{k,l}=\frac{\langle||f_k(x)-f_k(\tau_l(x))||^2\rangle_{x,\tau_l}}{\langle||f_k(x_1)-f_k(x_2))||^2\rangle_{x_1,x_2}}.
    \label{eq:d_kl}
\end{equation}
The measure Eq. \ref{eq:d_2} specializes Eq. \ref{eq:s_kl} to $k=2$ and $l=1$, pictured in \autoref{fig:tasksempos_lcn_s2} (B). T

he behaviours of $S_{k,l}$ and $D_{k,l}$ with respect to $P$ are shown for different values of $s=L=2$ in \autoref{fig:sens_lcn_L_2_s_2}. We vary $k$ and $l$ in $\{1,...,L\}$, and we analyse the sensitivity of the output too. For $k\geq l+1$ the sensitivities become significantly lower for large $P$. We checked the robustness of such observation for different values of $s$ and $L$ (not shown here). 
 
\section{Sample complexities and learnt representations for CNNs}
\label{app:sens_testerror_cnn}

\textbf{Architectures.} Shown in \autoref{fig:nets} (b). All layers consist of 512 channels.

\textbf{Training scheme.} Same as LCN, except for the learning rate: $lr = 0.01$ for $s=2$ and $s_0<4$, $lr = 0.01$ for $s=2$ and $s_0\geq 4$ and $lr = 0.003$ for $s\geq 2$. 

\begin{figure}[h]
    \centering
    \includegraphics[width=0.35\textwidth]{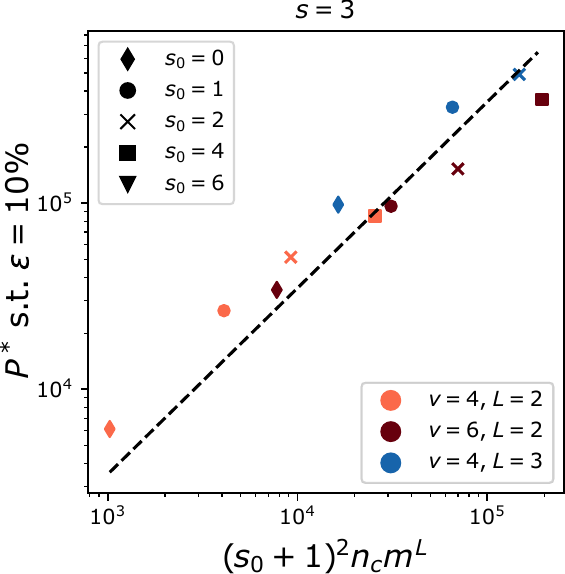}
    \caption{\textbf{CNN:} Empirical sample complexity $P^*$ to reach a $10\%$ test error $\varepsilon$ versus prediction Eq. \ref{eq:pstar_CNN} for $s=3$, for vocabulary size $v$ and different depths $L$ (different colors), maximal $m=v^{s-1}$ and different $s_0$ (different markers). }
    \label{fig:task_cnn_s3}
\end{figure}

\begin{figure}[h]
    \centering
    \includegraphics[width=.7\textwidth]{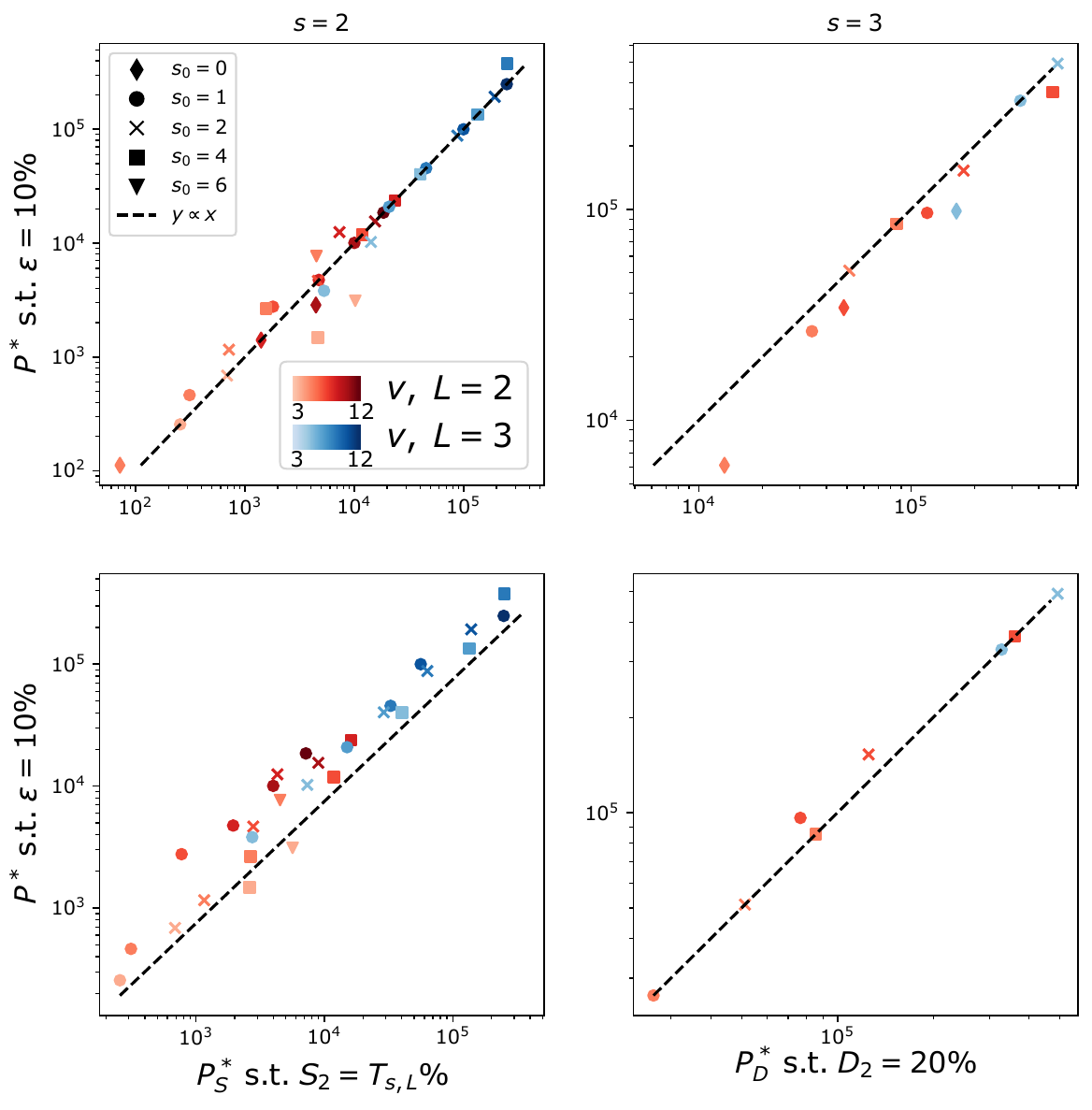}
    \caption{\textbf{CNN.} Left panels: empirical sample complexity $P^*_S$ to reach $S_{2}=T_{s,L}\%$ (with the threshold $T_{s,L} =40$ if $L=2$ and $s=2$, $50$ if $L=2$ and $s=3$, $20$ if $L=3$) versus empirical sample complexity $P^*$ to reach a $10\%$ test error $\varepsilon$ for (Top) $s=2$ and (Bottom) $s=3$, for different depths $L$ (red for $L=2$, blue for $L=3$) vocabulary sizes $v$ (different darkness), maximal $m=v^{s-1}$ and different $s_0$ (different markers). Right panels: same as left panels, but with the sample complexity $P^*_D$ to reach $D_{2}=20\%$. Both $P^*_S$ and $P^*_D$ align with $P^*$. The thresholds have been tuned based on the form of $S_{2}$ and $D_{2}$ versus $P$. }
    \label{fig:sem_pos_cnn}
\end{figure}

\textbf{Sample complexity figures.} In \autoref{fig:task_cnn_s3}, realised with patch size $s=3$, we show additional support to the scaling of the sample complexity Eq. \ref{eq:pstar_LCN}. In \autoref{fig:sem_pos_cnn} we show that, as for the LCNs, also for CNNs the insensitivity to diffeomorphism and to synonyms exchange are achieved at the same sample complexity $P^*_{CNN}$ at which the task is learned, defined in Eq. \ref{eq:pstar_CNN}.


\begin{figure}
    \centering
    \includegraphics[width=\textwidth]{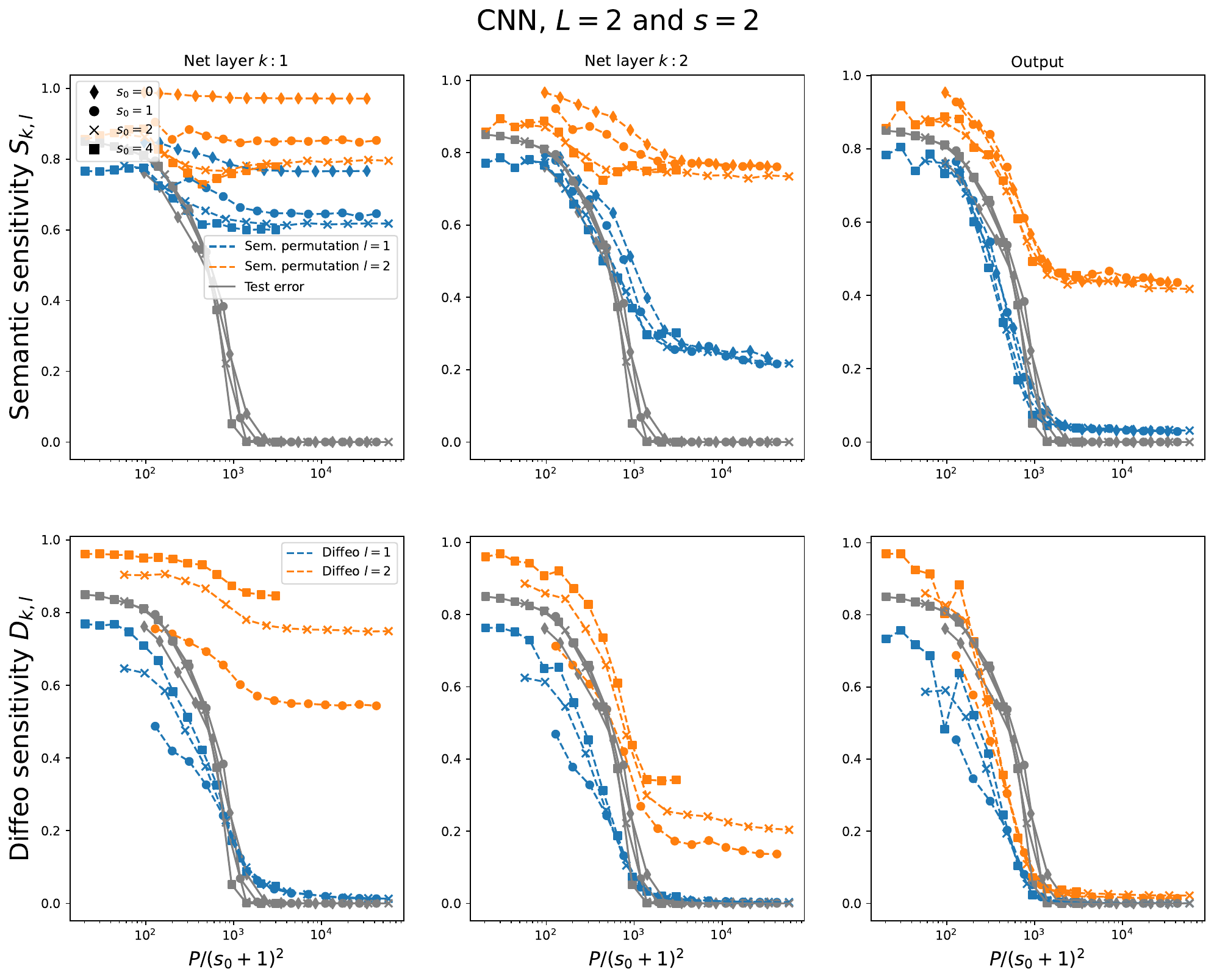}
    \caption{Top: sensitivity to synonyms exchange $S_{k,l}$ of a $L-$layer CNN, trained on the SRHM with $L=2$, $s=2$ and $m=v=8$, versus number of training points $P$ rescaled by prediction Eq. \ref{eq:pstar_CNN}. Going from left to right column the network layer $k$ increases from the layer closest to the input to the output. For each column at fixed $k$ there are plotted in color the sensitivities $S_{k,l}$ to synonyms exchange at the data level $l\in[0,...,L]$, and in grey the test error. Different markers stand for different $s_0$. Bottom: same as top, but with the sensitivity to diffeomorphisms $D_{k,l}$. The sensitivities $S_2$ and $D_2$ defined in Eq. \ref{eq:s_2} and Eq. \ref{eq:d_2} relate to the case $k=2$ and $l=1$ here.}
    \label{fig:sens_cnn_L_2_s_2}
\end{figure}

\textbf{Learnt representation figures.} In \autoref{fig:sens_cnn_L_2_s_2} we report the test error and the sensitivities to input transformations $S_{k,l}$ and $D_{k,l}$ defined in Eq. \ref{eq:s_kl} and Eq. \ref{eq:d_kl}, with respect to $P$, for CNNs trained on the SRHM for different values of $L$ and $s$. We vary $k$ and $l$ in $\{1,...,L\}$, and we analyse the sensitivity of the output too. As for the LCNs, for $k\geq l+1$ the sensitivities become significantly lower for large $P$. We checked the robustness of such observation for different values of $s$ and $L$ (not shown here).


\section{Sample complexities for FCNs}
\label{app:fcn}

\textbf{Architectures.} The networks are made by stacking $L$ full-connected layers, followed by a readout layer. All layers consist of 512 channels for $L=2$ and 256 channels for $L=3$.

\textbf{Training scheme.} Same as LCN, except for the learning rate: $lr = 0.01$ for $L=2$ and $lr = 0.003$ for $L=3$. 

\begin{figure*}[h]
    \centering
    \includegraphics[width=.65\textwidth]{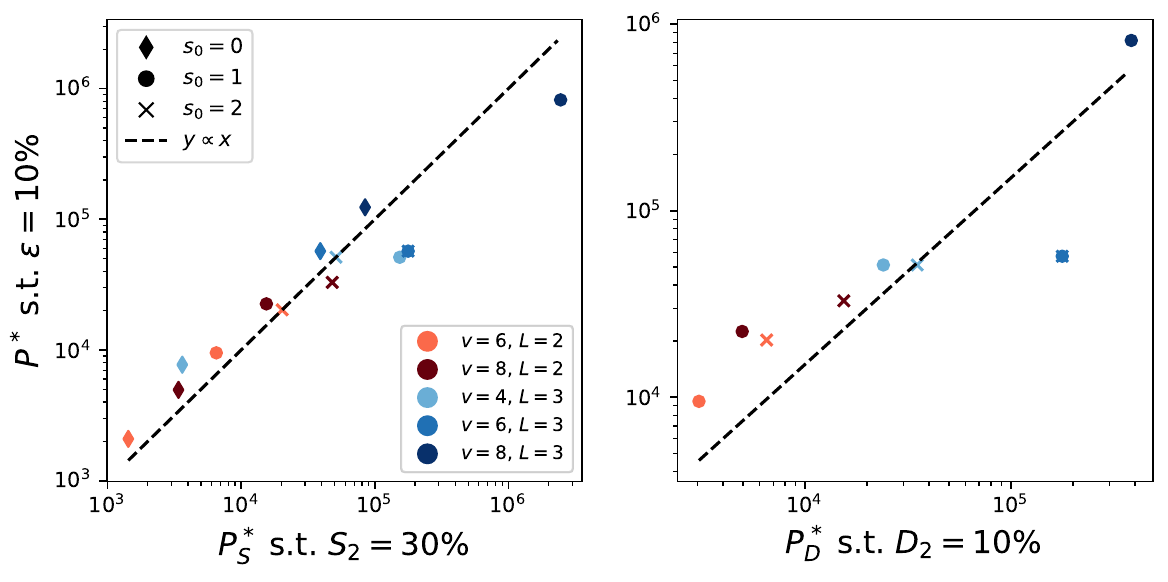}
    \caption{\textbf{FCN.} Left panel: empirical sample complexity $P^*$ to reach a $10\%$ test error $\varepsilon$ versus empirical sample complexity $P^*_S$ to reach $S_{2}=30\%$ for $s=2$, different vocabulary sizes $v$ and depths $L$ (different colors), number of classes $n_c=v$, maximal $m=v^{s-1}$ and different $s_0$ (different markers). 
    Right panel: same as left, for empirical sample complexity $P^*$ to reach a $10\%$ test error $\varepsilon$ versus empirical sample complexity $P^*_D$ to reach $D_{2}=10\%$.}
   
    \label{fig:fcn}
\end{figure*}

\textbf{Sample complexity figures.} In \autoref{fig:fcn}, realised with patch size $s=2$, we show that, as for the LCNs and CNNs, also for full-connected networks (FCN) the insensitivity to diffeomorphism and to synonyms exchange are achieved at the same sample complexity at which the task is learned.

\section{Correlation between test error and internal sensitivities}
\label{app:corr_int}

\begin{figure*}[h]
    \centering
    \includegraphics[width=.65\textwidth]{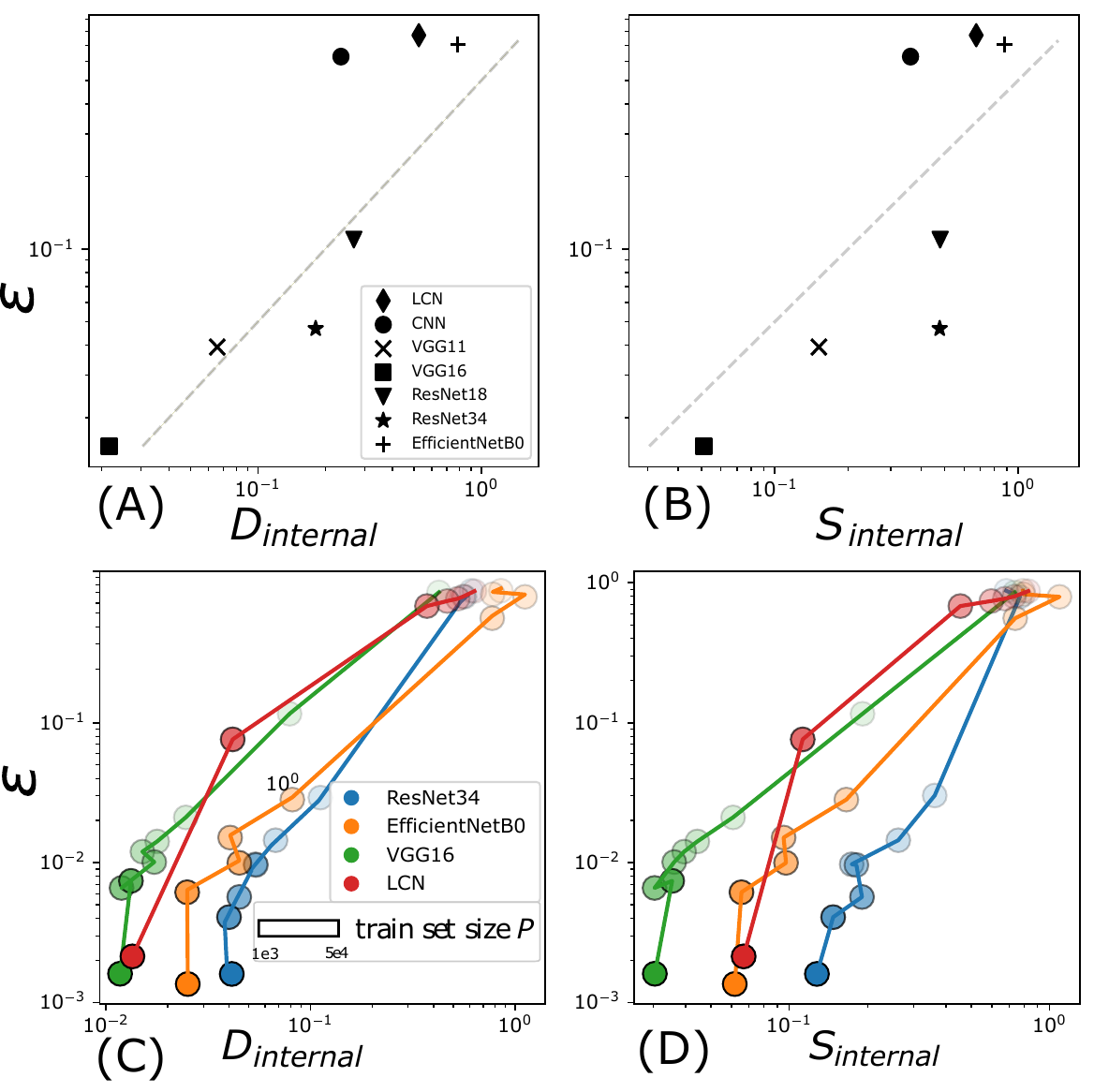}
    \caption{(A) Test error vs sensitivity $D_{\text{internal}}$ to diffeomorphisms of an internal deep layer for a CNN trained with $P=7400$  on the SHRM model, with parameters $L=s=s_0=2$ and $n_c=m=10$. \textcolor{black}{$D_{\text{internal}}$ is defined as the change of the hidden representation induced by a diffeomorphism applied on the input, see Eq. \ref{eq:d_2}.} The internal layer is at 80\% relative depth of the architecture, except for the 2 hidden-layer LCN, where it corresponds to the second layer. The test error shows a remarkable correlation with the sensitivities. A grey line, corresponding to a power-law, guides the eye. For details about the architectures and their training process, see \autoref{app:sens_testerror_newnets}.  (B) Same as (A)  for sensitivity $S_{\text{internal}}$ to synonymic exchanges, defined as the change of the hidden representation induced by an exchange of synonyms applied on the input, see Eq. \ref{eq:s_2}. (C) and (D): as top panels (A) and (B), for increasing $P$ (increasing opacity). The sensitivities of the network output yield the same observations, as shown in \autoref{fig:phasediagram_new}. }
   
    \label{fig:phasediagram_new_int}
\end{figure*}




\end{document}